\documentclass[conference]{IEEEtran}
\IEEEoverridecommandlockouts
\usepackage{cite}
\usepackage{amsmath,amssymb,amsfonts}
\usepackage{algorithmic}
\usepackage{graphicx}
\usepackage{textcomp}
\usepackage{xcolor}
\usepackage{multirow}
\usepackage{booktabs}
\def\BibTeX{{\rm B\kern-.05em{\sc i\kern-.025em b}\kern-.08em
    T\kern-.1667em\lower.7ex\hbox{E}\kern-.125emX}}
\makeatletter

\newcommand{\linebreakand}{%
\end{@IEEEauthorhalign}
\hfill\mbox{}\par
\mbox{}\hfill\begin{@IEEEauthorhalign}
}
\makeatother

\begin{document}

\title{Histogram-guided Video Colorization Structure with Spatial-Temporal Connection\\
\thanks{
This work is supported by Natural Science Foundation of China under Grant No. 62202429, U20A20196 and Zhejiang Provincial Natural Science Foundation of China under Grant No. LY23F020024, LR21F020002.}
\thanks{$*$ The corresponding author.}
}

\author{
\IEEEauthorblockN{Zheyuan Liu}
\IEEEauthorblockA{\textit{College of Computer Science and Technology} \\
\textit{Zhejiang University of Technology}\\
Hangzhou, China \\
zheyuanliu@zjut.edu.cn}
\and
\IEEEauthorblockN{Pan Mu$^*$}
\IEEEauthorblockA{\textit{College of Computer Science and Technology} \\
\textit{Zhejiang University of Technology}\\
Hangzhou, China \\
panmu@zjut.edu.cn}
\linebreakand
\IEEEauthorblockN{Hanning Xu}
\IEEEauthorblockA{\textit{College of Computer Science and Technology} \\
\textit{Zhejiang University of Technology}\\
Hangzhou, China \\
202006010425@zjut.edu.cn}
\and
\IEEEauthorblockN{Cong Bai}
\IEEEauthorblockA{\textit{College of Computer Science and Technology} \\
\textit{Zhejiang University of Technology}\\
Hangzhou, China \\
congbai@zjut.edu.cn}
}

\maketitle

\begin{abstract}
Video colorization, aiming at obtaining colorful and plausible results from grayish frames, has aroused a lot of interest recently. Nevertheless, how to maintain temporal consistency while keeping the quality of colorized results remains challenging. To tackle the above problems, we present a Histogram-guided Video Colorization with Spatial-Temporal connection structure (named ST-HVC). To fully exploit the chroma and motion information, the joint flow and histogram module is tailored to integrate the histogram and flow features. To manage the blurred and artifact, we design a combination scheme attending to temporal detail and flow feature combination. We further recombine the histogram, flow and sharpness features via a U-shape network. Extensive comparisons are conducted with several state-of-the-art image and video-based methods, demonstrating that the developed method achieves excellent performance both quantitatively and qualitatively in two video datasets. 
\end{abstract}


\begin{figure*}[t]
	\centering 
	\includegraphics[width=1\textwidth]{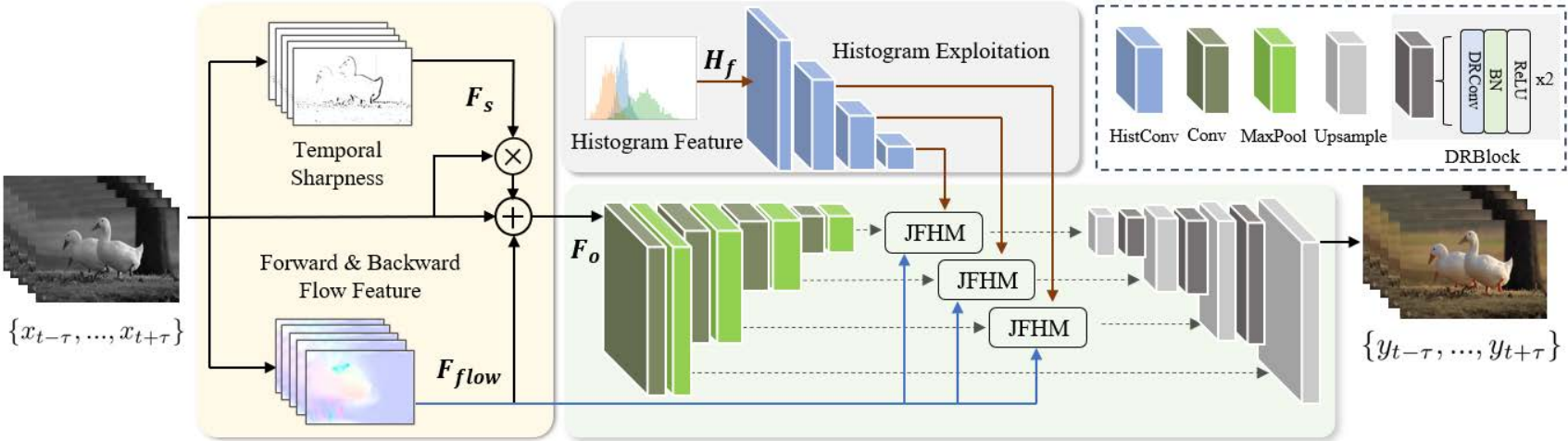}
	\caption{Model overview of the proposed ST-HVC. Temporal sharpness and flow features are computed and forwarded into U-shape network with the frames. Histogram and flow features are integrated into joint flow and histogram feature module (short for JFHM) to facilitate the colorization process. }
	\label{fig:overview} 
\end{figure*}

\section{Introduction}
Video colorization task aims to convert the gray frame sequences into colorful and plausible ones, which has wide applications in domains like old films restoration and anime creation. Video colorization can also assist other tasks like video action recognition, detection, tracking and segmentation. 

Although significant progress has been achieved, automatic video colorization still poses the following three challenges: the chroma quality of produced images, the temporal consistency between frames and the possibility of producing distinct colorization result. 
Image-based colorization solutions~\cite{vitoria2020chromagan,weng2022ct2,kumar2021colorization,zhang2022histogram,zhang2016colorful,zhang2017real,wu2021towards} can achieve satisfactory visual result.
Nonetheless, those methods tend to take longer time to colorize images and the temporal coherence of generated results is relatively poor.
Video-based methods~\cite{lai2018learning,liu2021temporally,lei2019fully,zhang2019deep,lei2020blind,zhao2022vcgan, mu2021triple} are aimed to produce colorized frames with vivid color and minimum temporal flickering. But to achieve a trade-off between quality of single frame and the temporal consistency of neighbor frames is hard.

The final challenge makes video colorization an ill-posed problem, since the same object can have distinct but possible colors at the same time (i.e., the color of T-shirt can be green or blue, a flower can be purple or white). To solve this problem, researchers either build models to  produce multiple results all at once, or tend to seek references to guide the model to colorize the video frames. 
User-guided colorization like~\cite{zhang2017real} seek the indication from users to guide the model.
Exemplar-based methods like~\cite{zhang2019deep,he2018deep,zhao2021color2embed} utilize exemplar images as color references. Thus the generated results differs with different exemplars. 
Histograms can also be served as color references to provide an global color distribution for models. For instance, HistoGAN~\cite{afifi2021histogan} employs histogram to manipulate the color of GAN-generated images, \cite{zhang2022histogram} uses histograms to facilitate the extraction of semantic information.  

Over the past several years, the attention mechanism along with the transformer architecture has been broadly used in vision based tasks like detection~\cite{carion2020end}, segmentation~\cite{zheng2021rethinking} and image restoration~\cite{liu2020investigating, mu2022structure}. 
In the field of colorization, ColTran~\cite{kumar2021colorization} proposes a transformer model based on the theory of probability  and samples colors from the learned distribution to generate diverse and plausible results. $CT^2$~\cite{weng2022ct2} converts the colorization task as a classification problem and introduces color tokens into their model while limiting the range of $Lab$ color space.

In this work, we develop a histogram-guided video colorization with spatial-temporal connection structure, along with a developed joint flow and histogram feature module, spatial-temporal connection scheme and a feature recombination-aimed U-shape network. 
In particular, the developed method firstly calculates the temporal sharpness and flow feature, to help the model better restore the blurred detail caused by motion and the artifact brought by optical flow. 
Aiming to integrate the flow and histogram information to facilitate the colorization, we present joint flow and histogram feature module and employ it in skip connections and in the bottleneck of the U-shape network. 
We extract the reference histogram feature in multiple level via splatting the pixel histogram along the range dimension.
We feed merely the histogram of the middle frame in a video into our model, which can further boost the application of the proposed method. 
We make the following contributions:
\begin{itemize}
	\item[$\bullet$] 
	We propose a novel histogram-guided video colorization network with spatial-temporal connection structure (i.e., ST-HVC) to tackle the color assignment and temporal consistency challenges.
	\item[$\bullet$] 
	For the first time we calculate the temporal sharpness and flow feature to form the inputs along with the frames. This design can better manage the blurred detail caused by motion and the artifact brought by flow.
	\item[$\bullet$] 
	To ameliorate the wrongly-assigned color and improve the temporal coefficient, we integrate the histogram of the middle frame and flow features into a joint module (i.e., JFHM) to guide the colorization process.
	\item[$\bullet$] 
	The experiments confirm that the proposed method significantly outshines existing video colorization approaches both quantitatively and qualitatively. 
\end{itemize}

\section{Method}
\subsection{Method Overview}
Video colorization task aims at attaining colorful and plausible results from black and white frames while reducing the flickering artifacts and maintain the quality of colorized images.
In Fig.~\ref{fig:overview}, we illustrate the overall pipeline of our developed network ST-HVC.
The proposed method takes in $2\tau+1$ video frames as input and then computes the temporal sharpness $F_s$ ~\cite{pan2020cascaded,tran2021explore} and flow feature $F_{flow}$, the two features are then concatenated with original frames $\{x_{t-\tau},...,x_{t+\tau}\}$ and are forwarded into a dynamic region convolution~\cite{chen2021dynamic} based encoder-decoder architecture. 
To remove the artifacts commonly occurred in video restoration results and boost the consistency between frames, four JFHMs are placed in skip connections and in bottleneck.
We decouple the original multi-head self-attention in JFHM along spatial and temporal dimension to better utilize the spatially histogram color hint and temporally flow feature.
We extract the histogram reference feature $H_f$ in multiple levels by introducing bilateral grid~\cite{chen2007real} and via splatting the pixel histogram along the range dimension.
It is noteworthy that the reference we feed into the model is merely the middle frame of the video. We don't need any other references to guide the colorization process.

\subsection{Temporal Detail and Flow Combining}\label{input}
Temporal sharpness and flow features are firstly computed according to the frames before they are forwarded into U-shape network. 
This prior is based on the observation that the same object in different frames has sharp and blurred pixels simultaneously. Since many colorization works cannot perform well in blurred details, the temporal sharpness can prompt our model to sense the sharpness and help to restore those details. 

The temporal sharpness $F_s$ calculated from frames  $\{x_{t-\tau},...,x_{t+\tau}\}$ 
indicates the sharpness area of frames. The overall input $F_o$ can be formed below:
\begin{equation}
	F_o= Concat(x_{i}, (F_s \otimes x_i), F_{flow}), i\in \left [t-\tau, t-\tau \right ]
\end{equation}
where $Concat$ and $\otimes$ represents the concatenation and multiplication. We use RAFT~\cite{teed2020raft} to form the optical flow $F_{flow}$.

\begin{figure}[t]
	\centering 
	\includegraphics[width=0.47\textwidth]{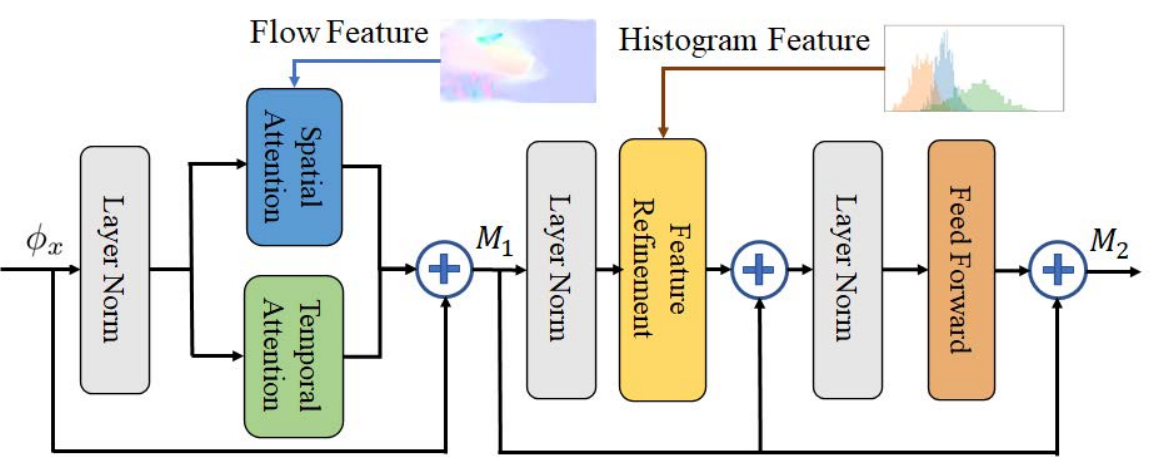}
	\caption{Structure of JFHM with flow and histogram feature integrated. $\phi_x$ stands for the input feature. $M_1$ means the middle feature map and $M_2$ denotes the output of JFHM module. }
	\label{fig:stahg} 
\end{figure}

\begin{figure*}[ht]
	\centering 
	\begin{tabular}{c@{\extracolsep{0.2em}}c@{\extracolsep{0.2em}}c@{\extracolsep{0.2em}}c@{\extracolsep{0.2em}}c@{\extracolsep{0.2em}}c@{\extracolsep{0.2em}}c@{\extracolsep{0.2em}}c@{\extracolsep{0.2em}}c}
		&\includegraphics[width=0.115\textwidth]{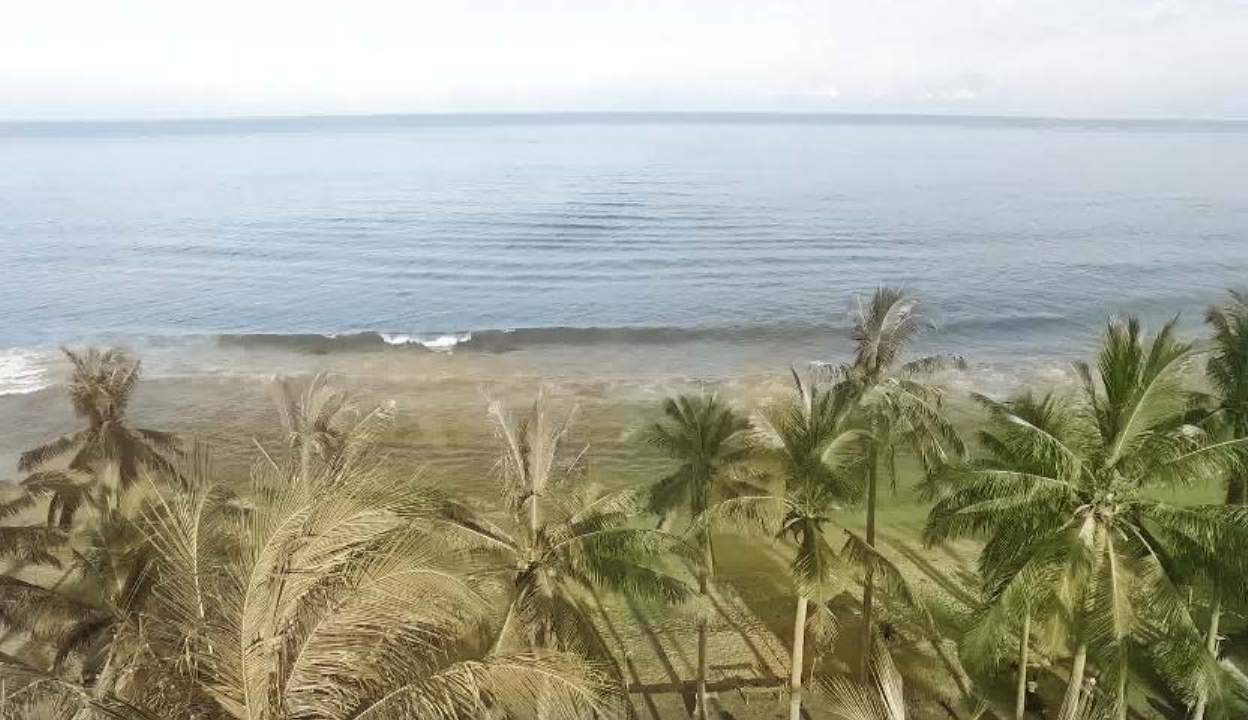}
		&\includegraphics[width=0.115\textwidth]{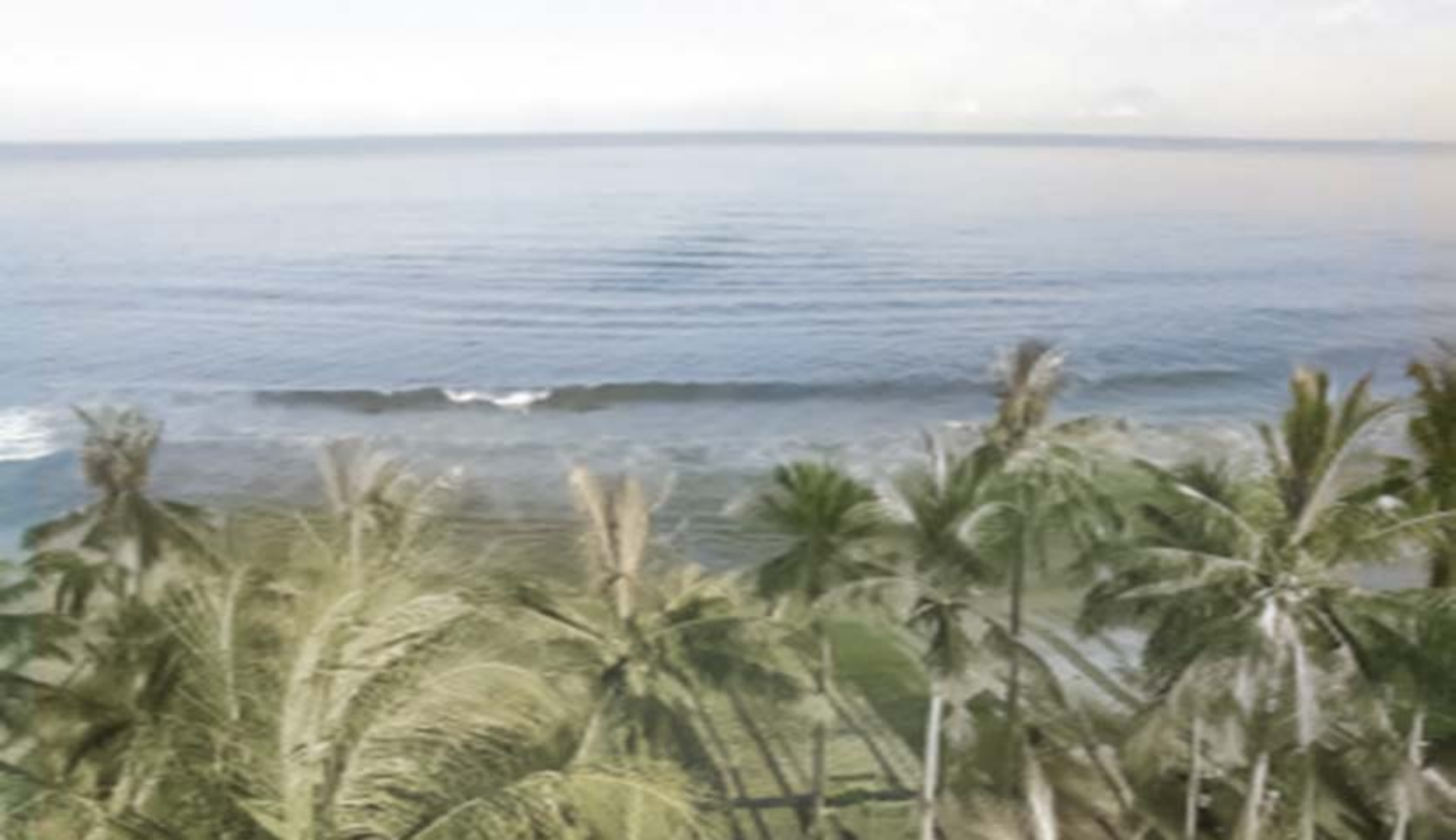}
		&\includegraphics[width=0.115\textwidth]{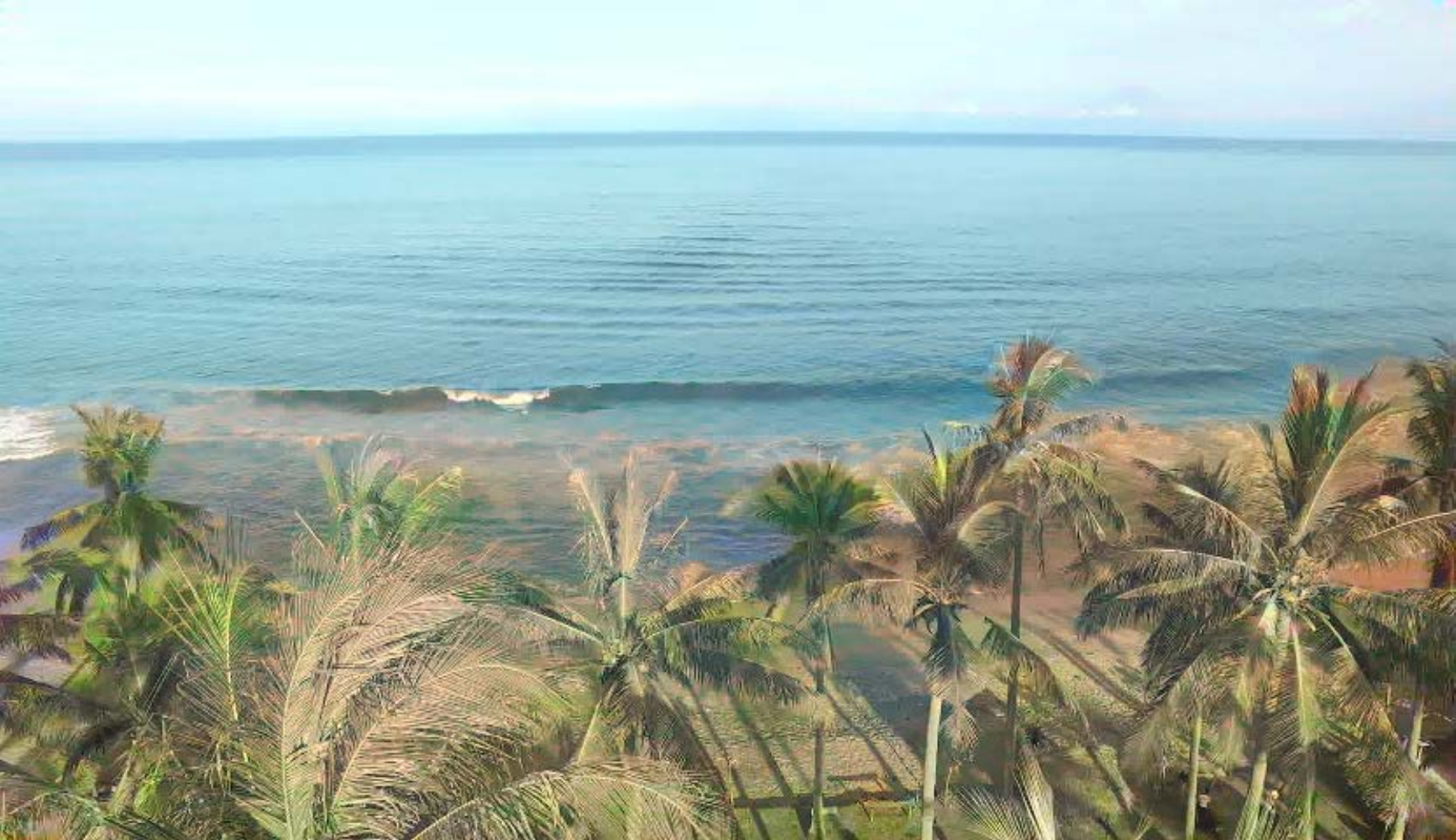}
		&\includegraphics[width=0.115\textwidth]{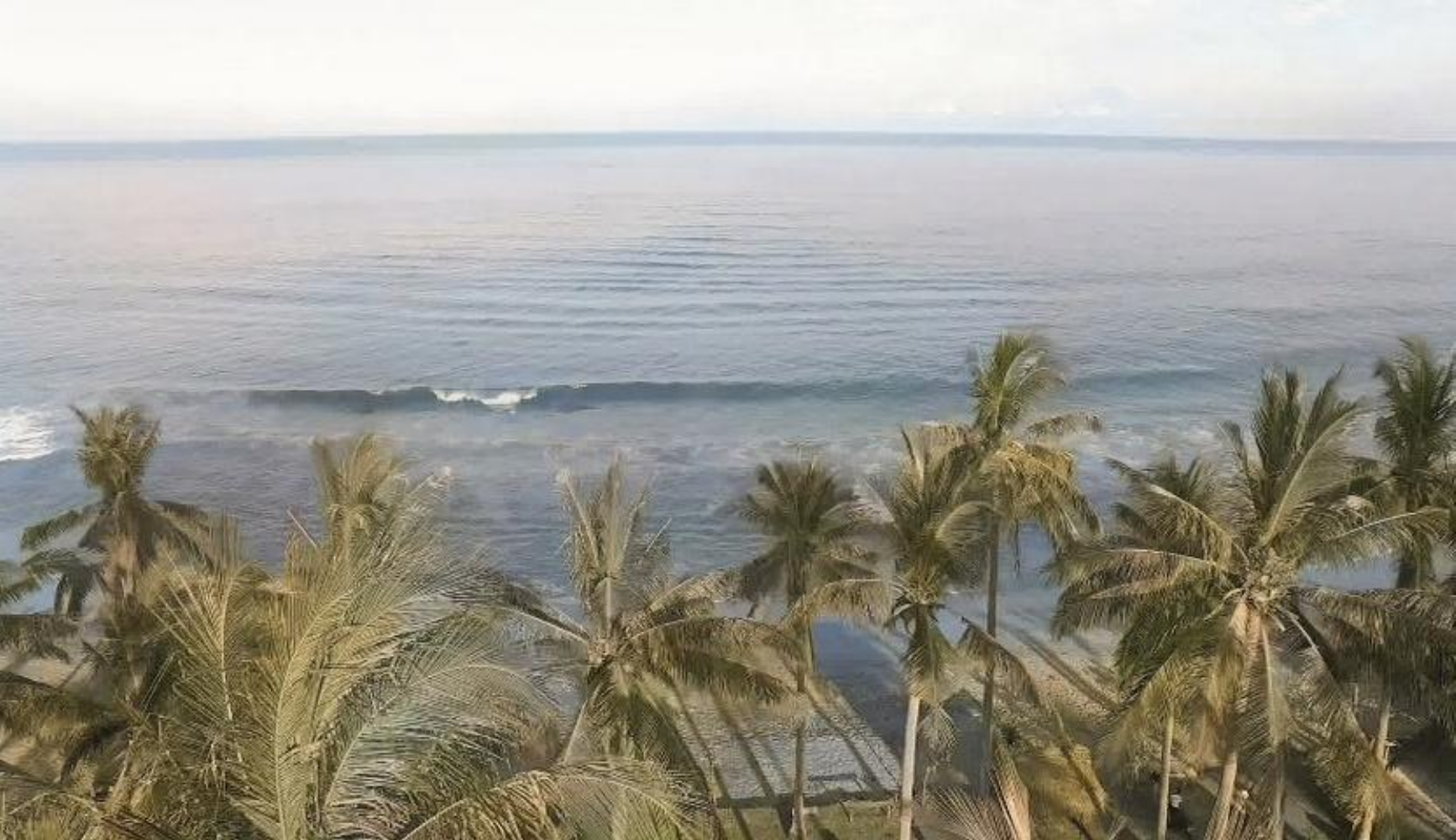}
		&\includegraphics[width=0.115\textwidth]{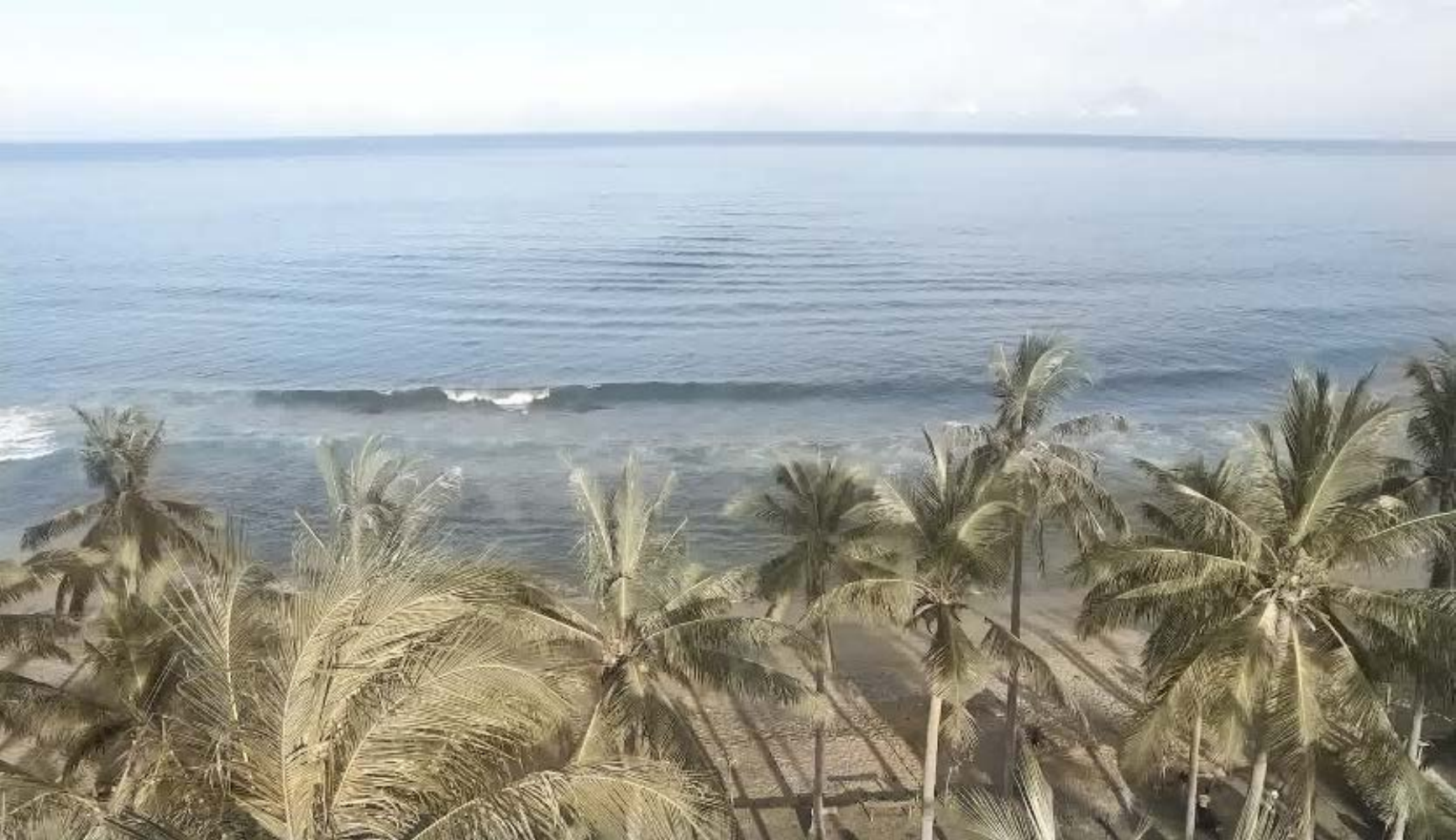}
		&\includegraphics[width=0.115\textwidth]{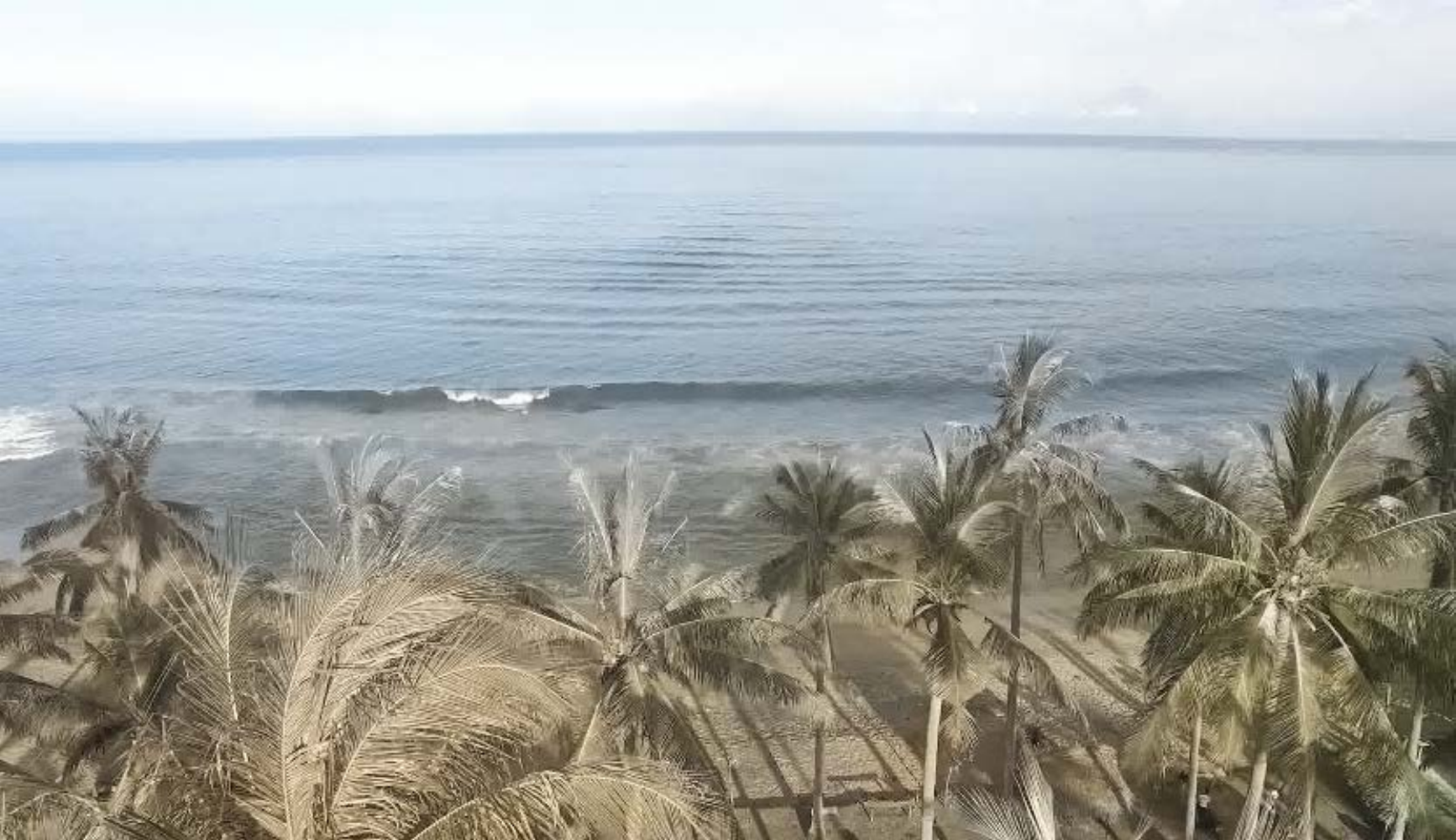}
		&\includegraphics[width=0.115\textwidth]{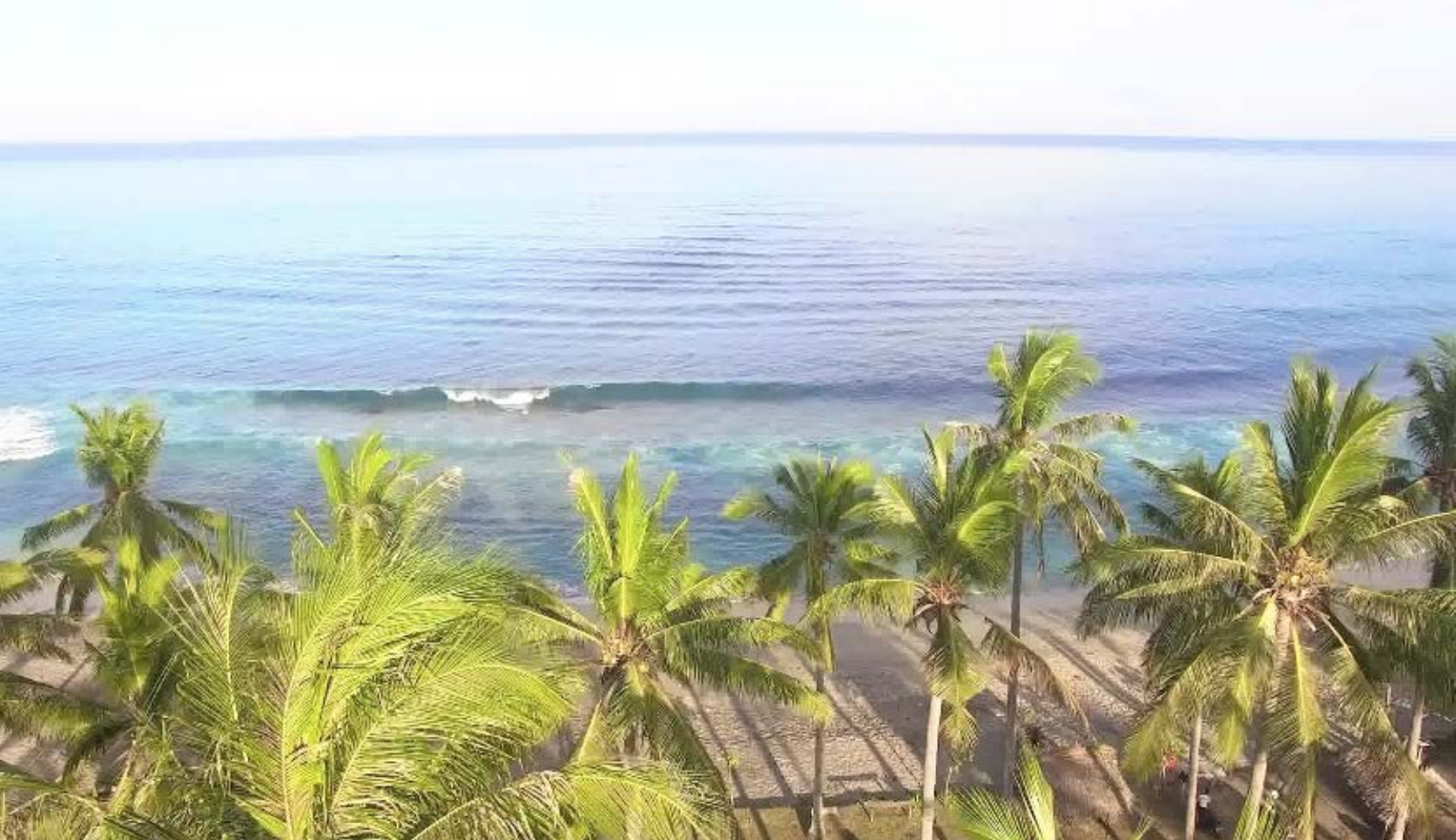}
		&\includegraphics[width=0.115\textwidth]{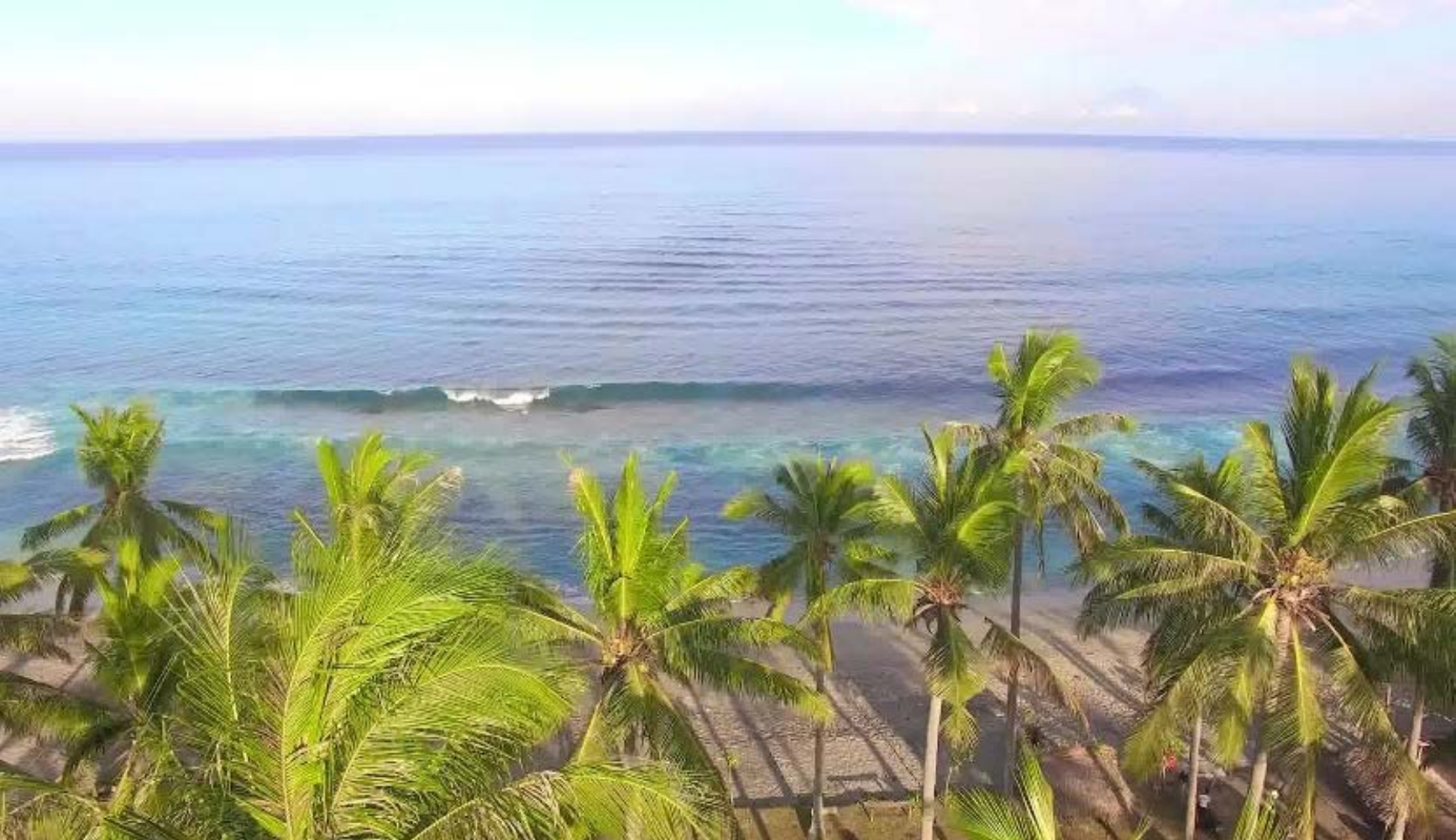}
		\\
		&\includegraphics[width=0.115\textwidth]{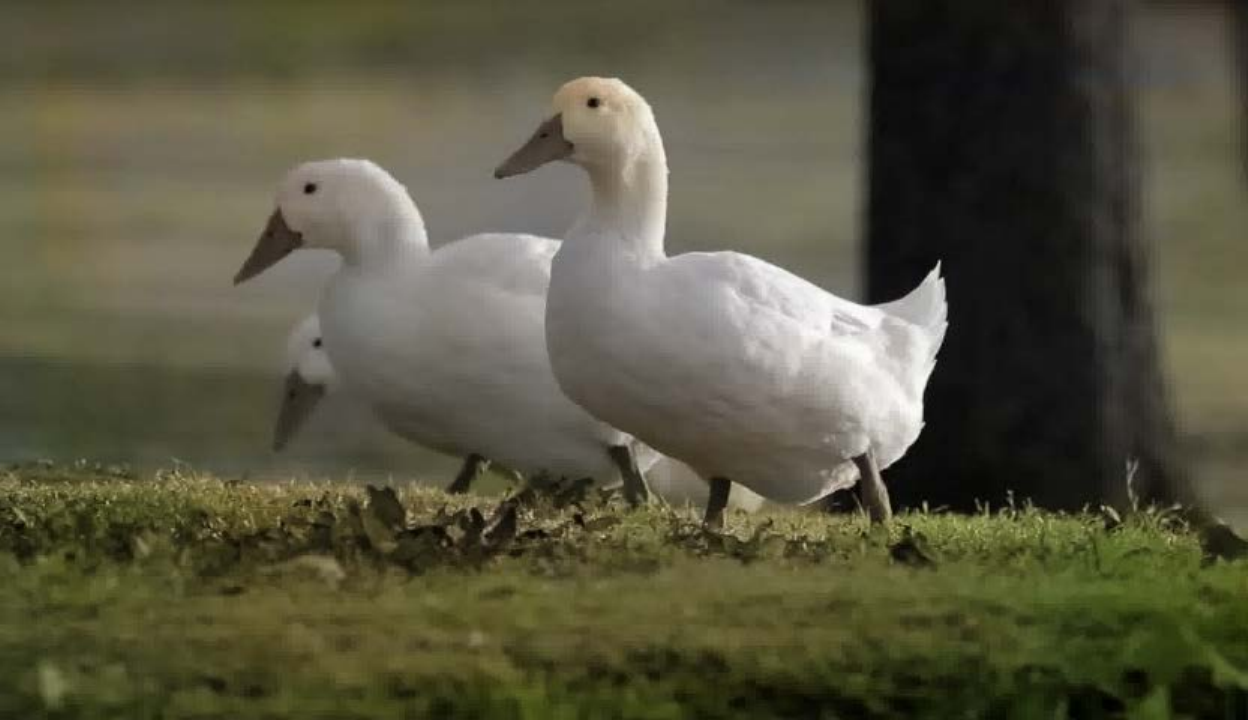}
		&\includegraphics[width=0.115\textwidth]{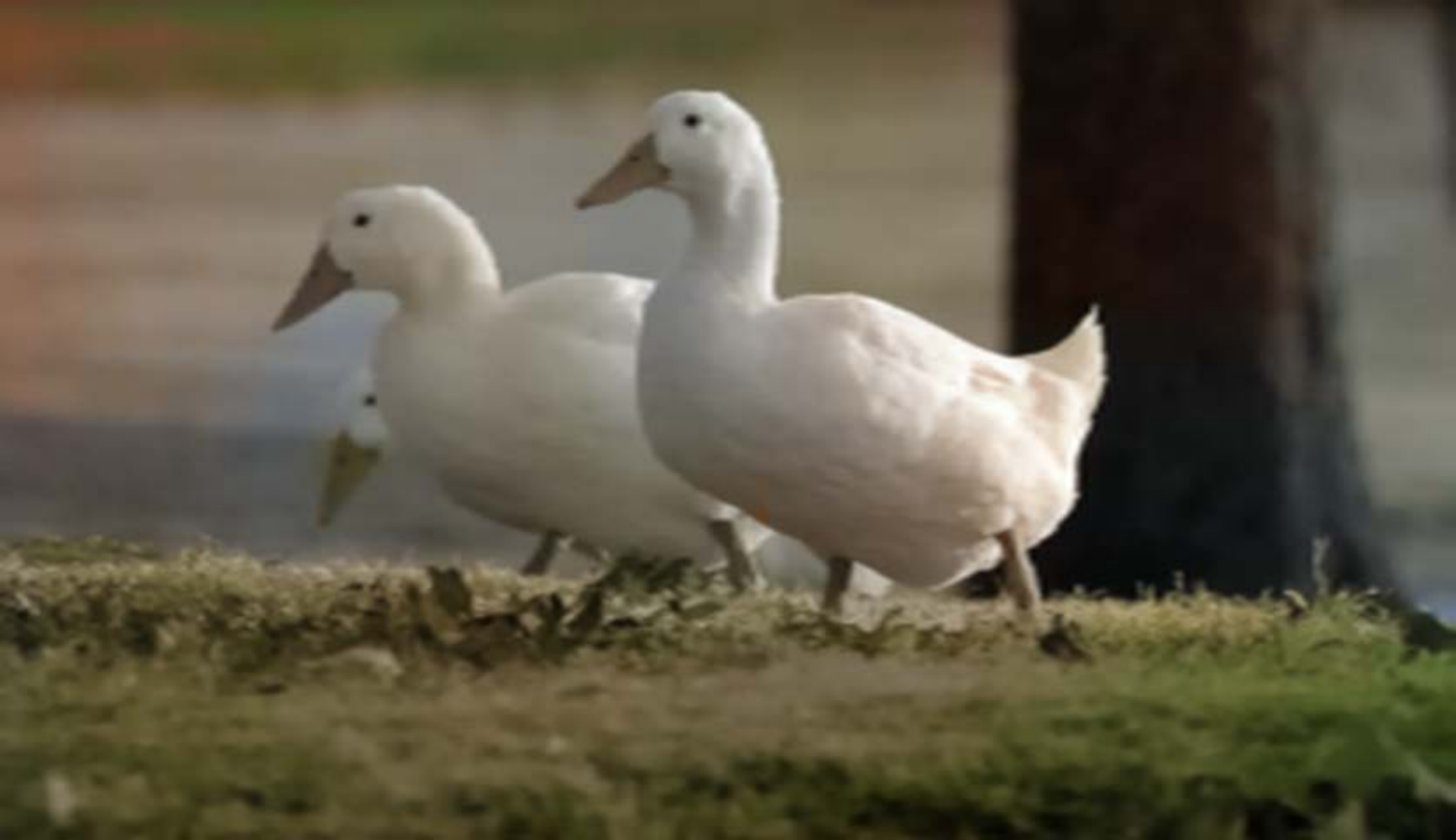}
		&\includegraphics[width=0.115\textwidth]{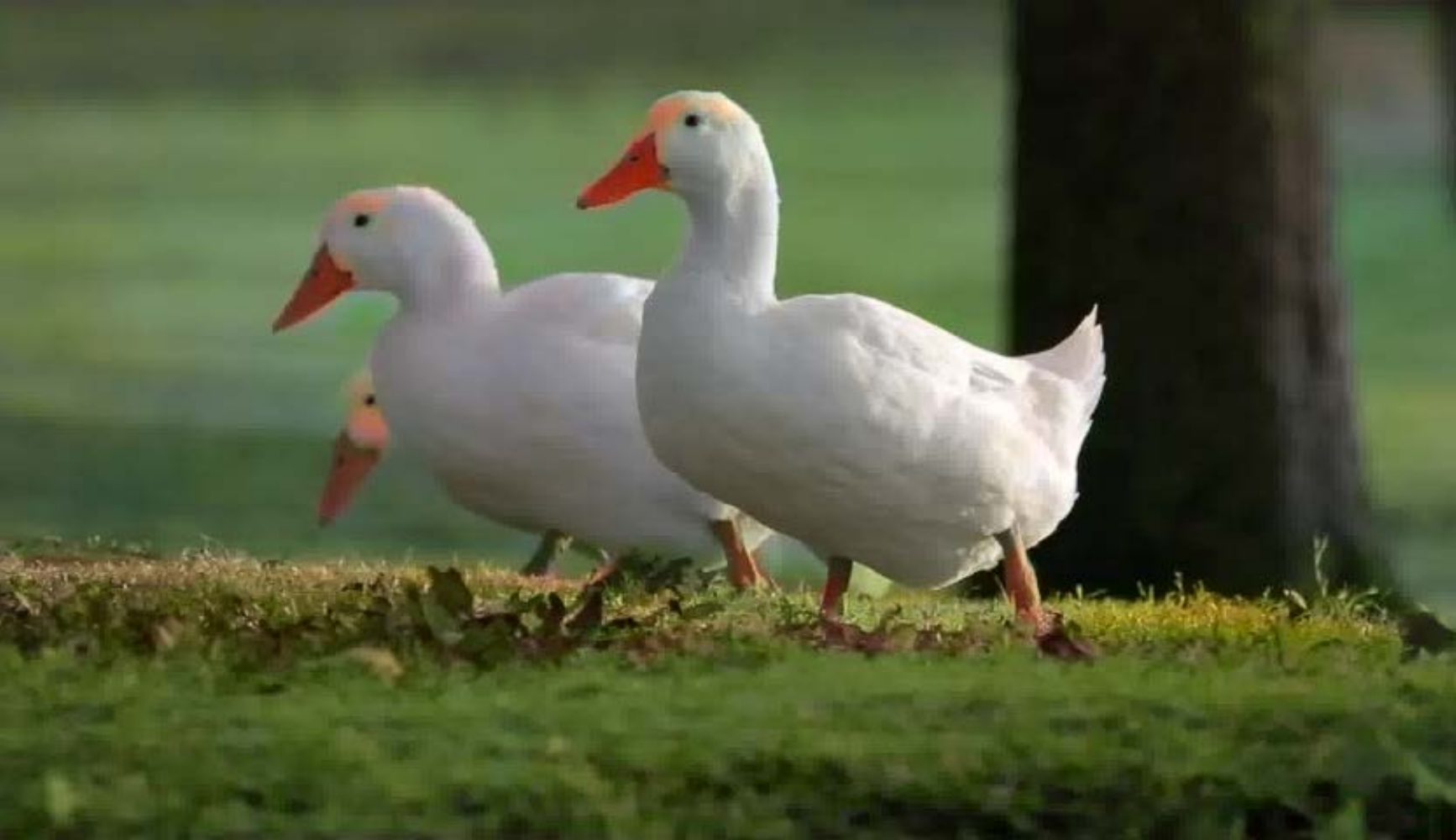}
		&\includegraphics[width=0.115\textwidth]{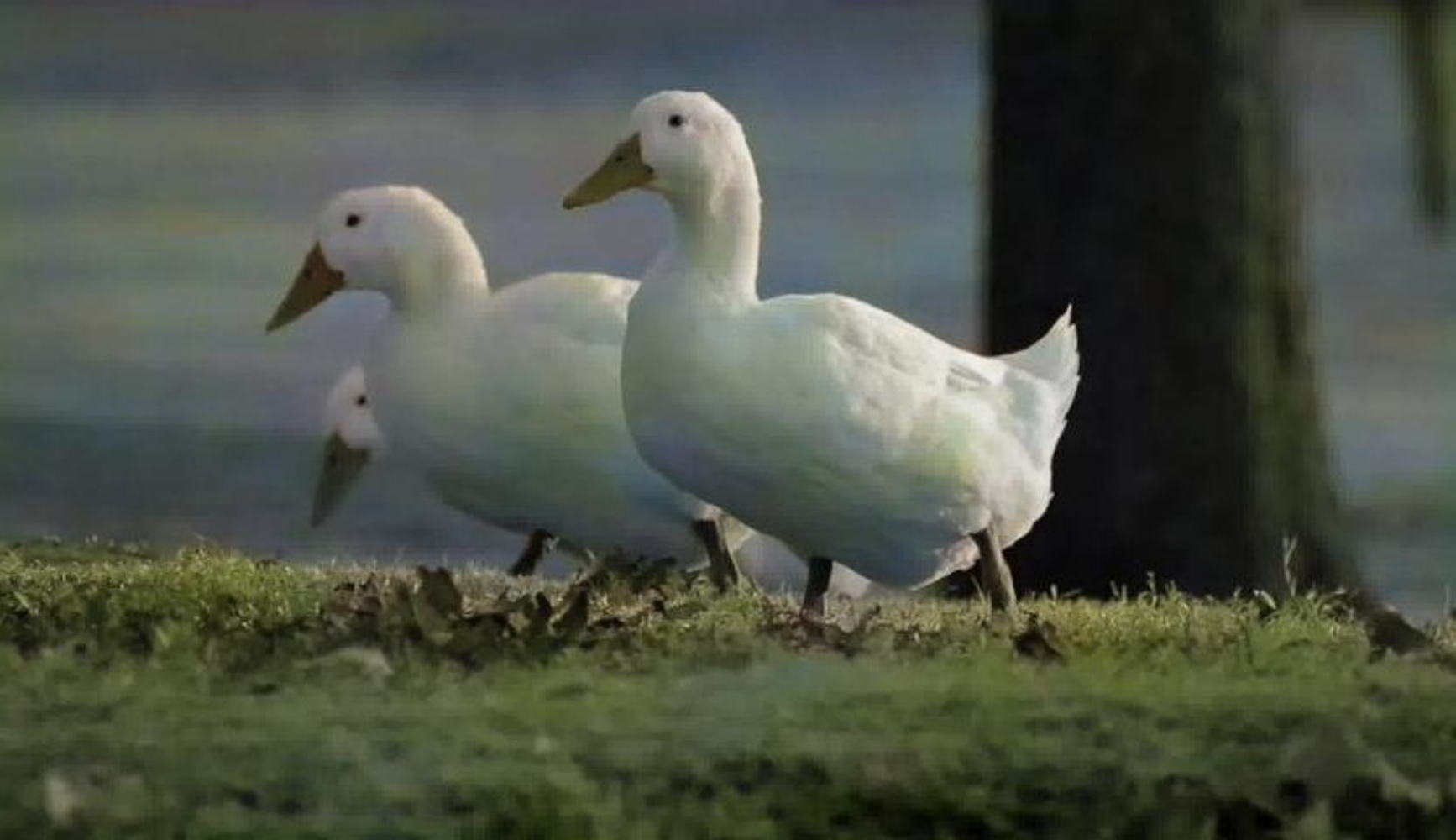}
		&\includegraphics[width=0.115\textwidth]{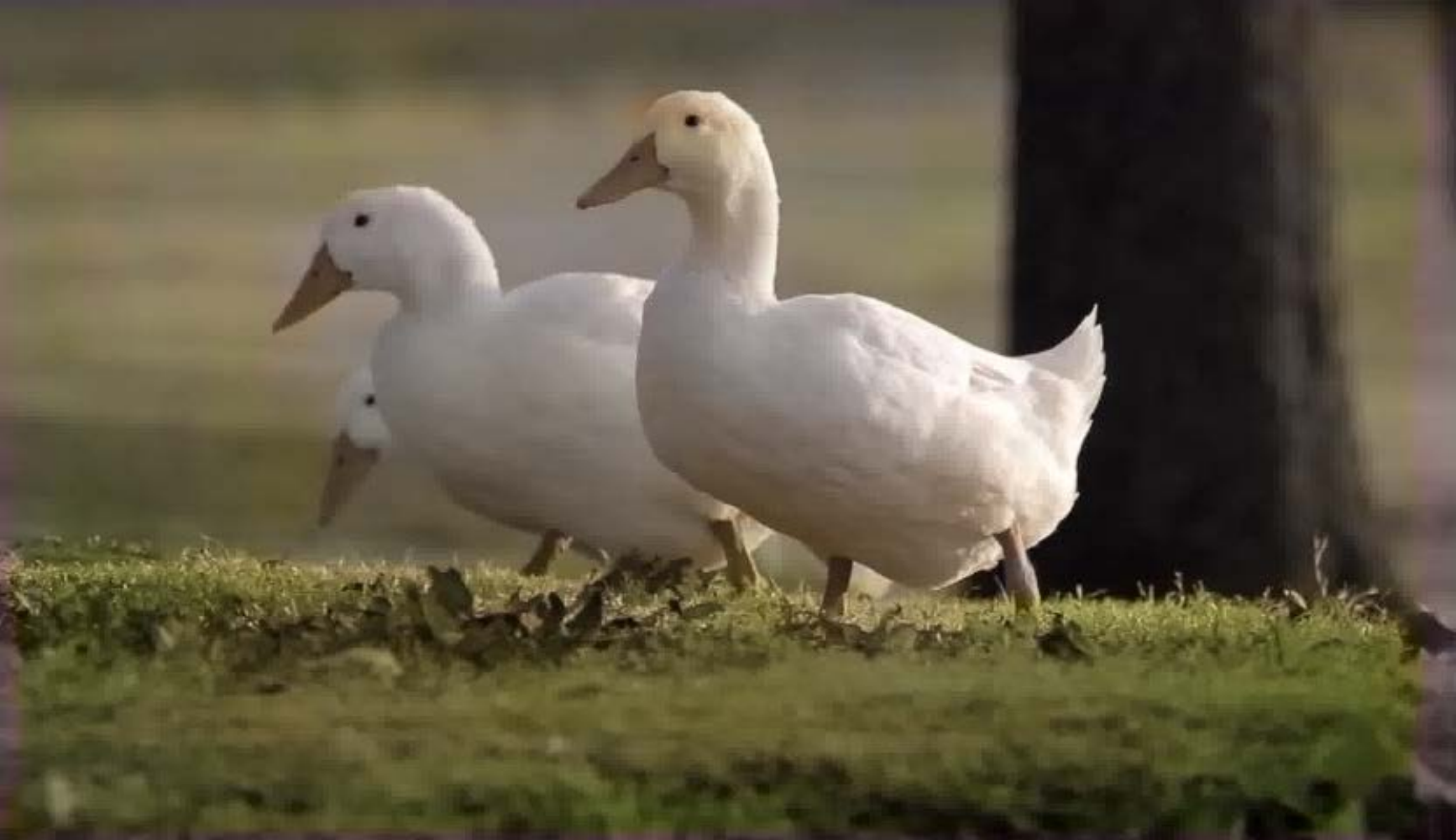}
		&\includegraphics[width=0.115\textwidth]{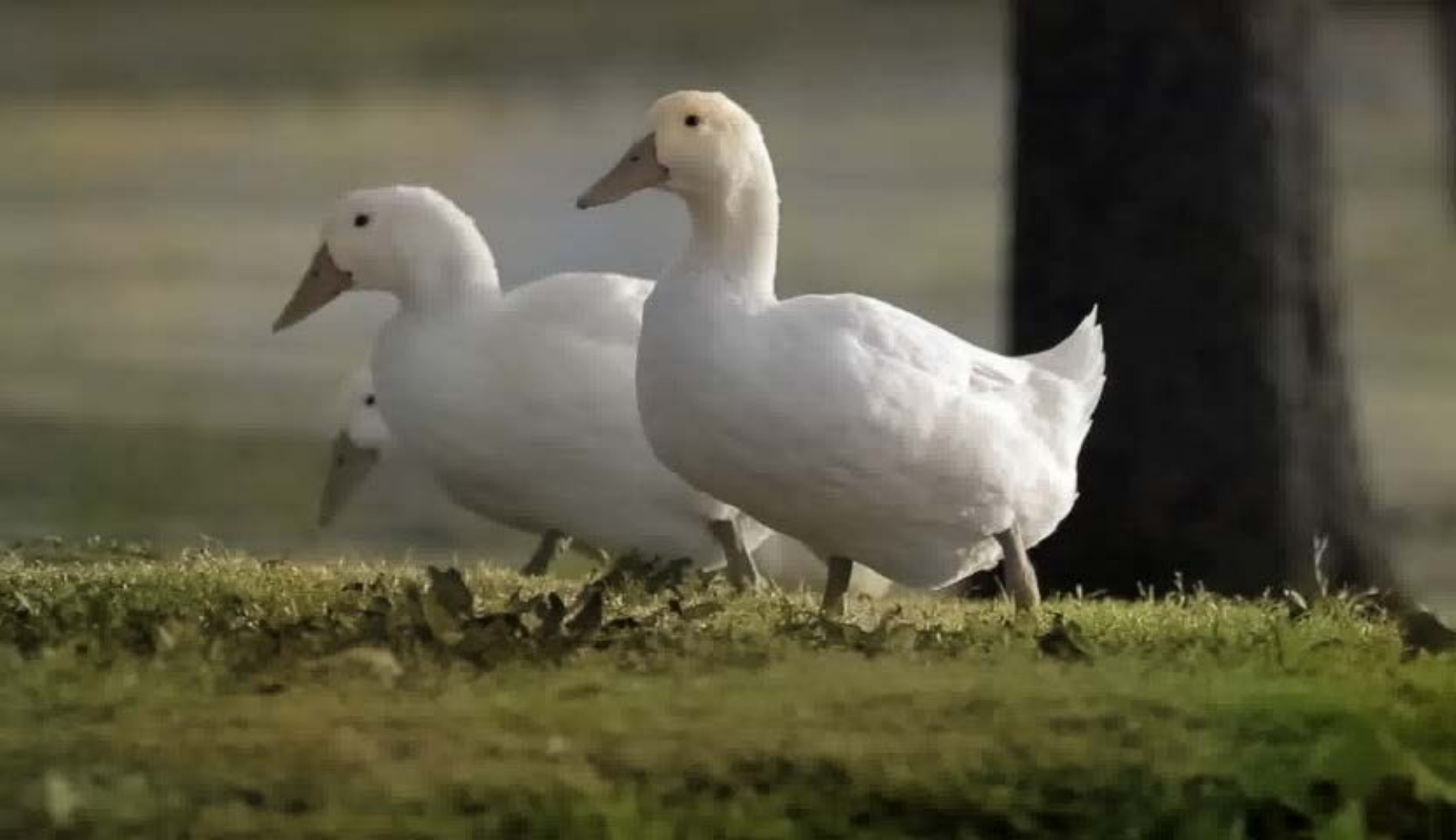}
		&\includegraphics[width=0.115\textwidth]{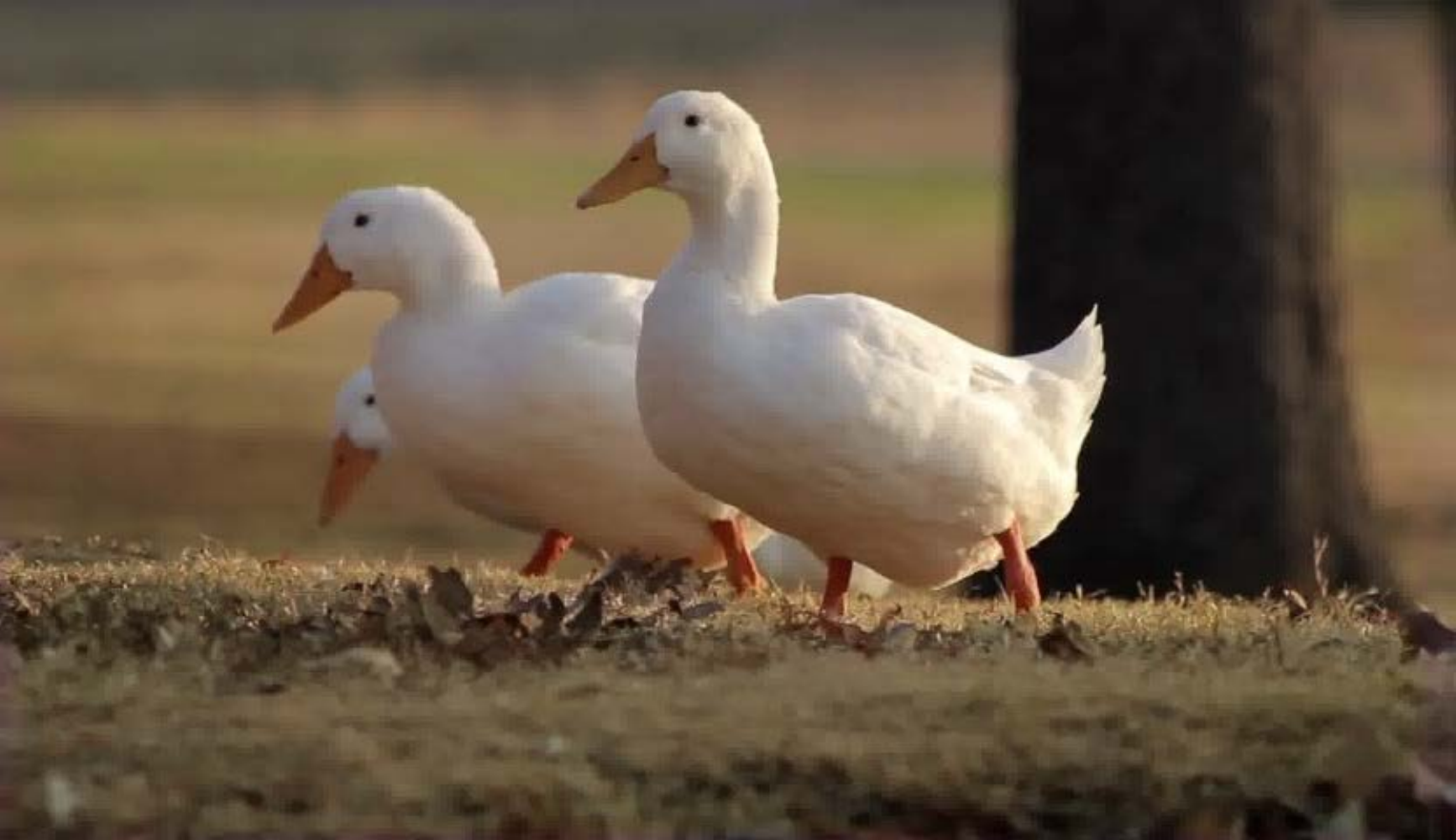}
		&\includegraphics[width=0.115\textwidth]{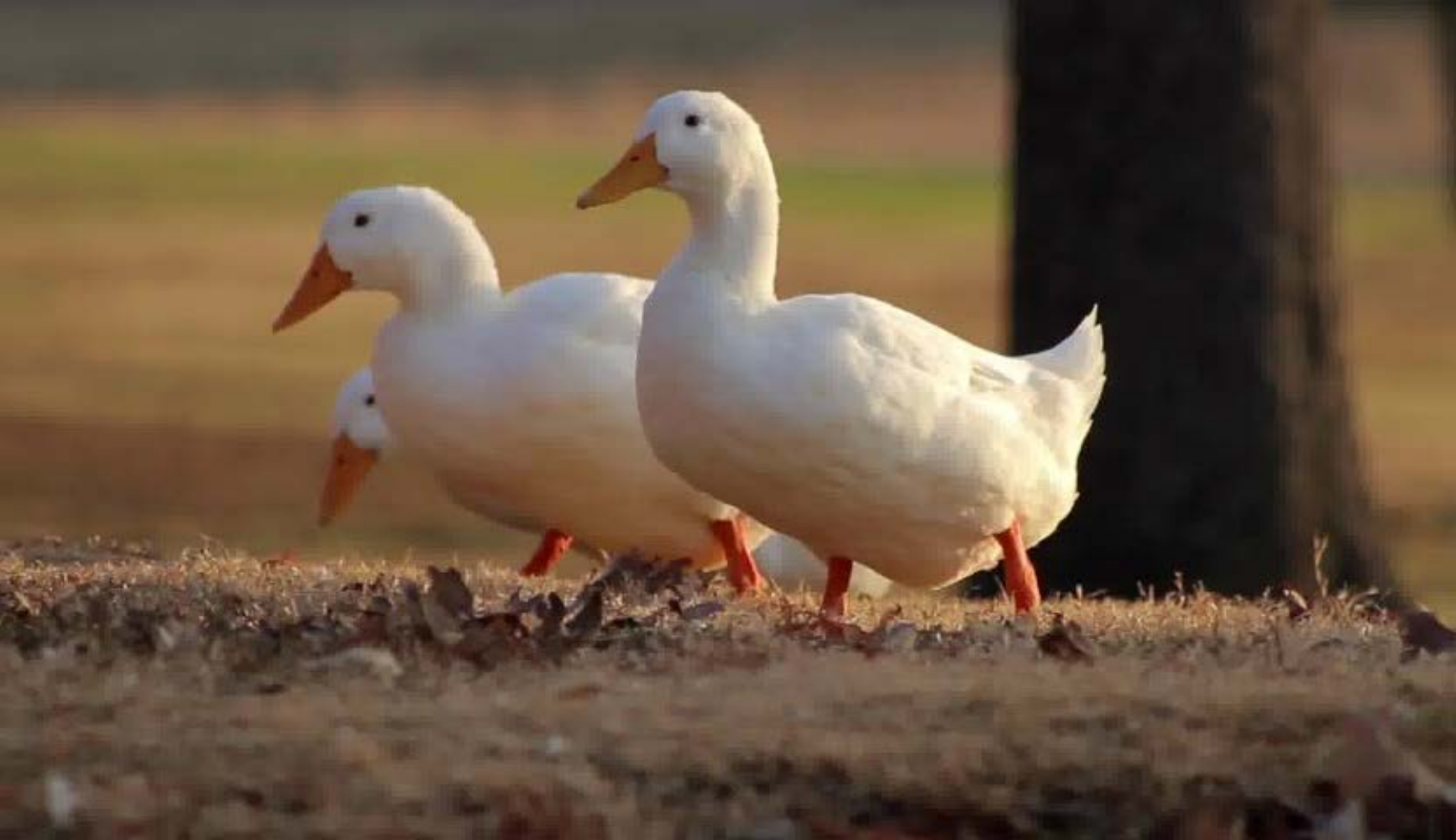}
		\\
		
		&\includegraphics[width=0.115\textwidth]{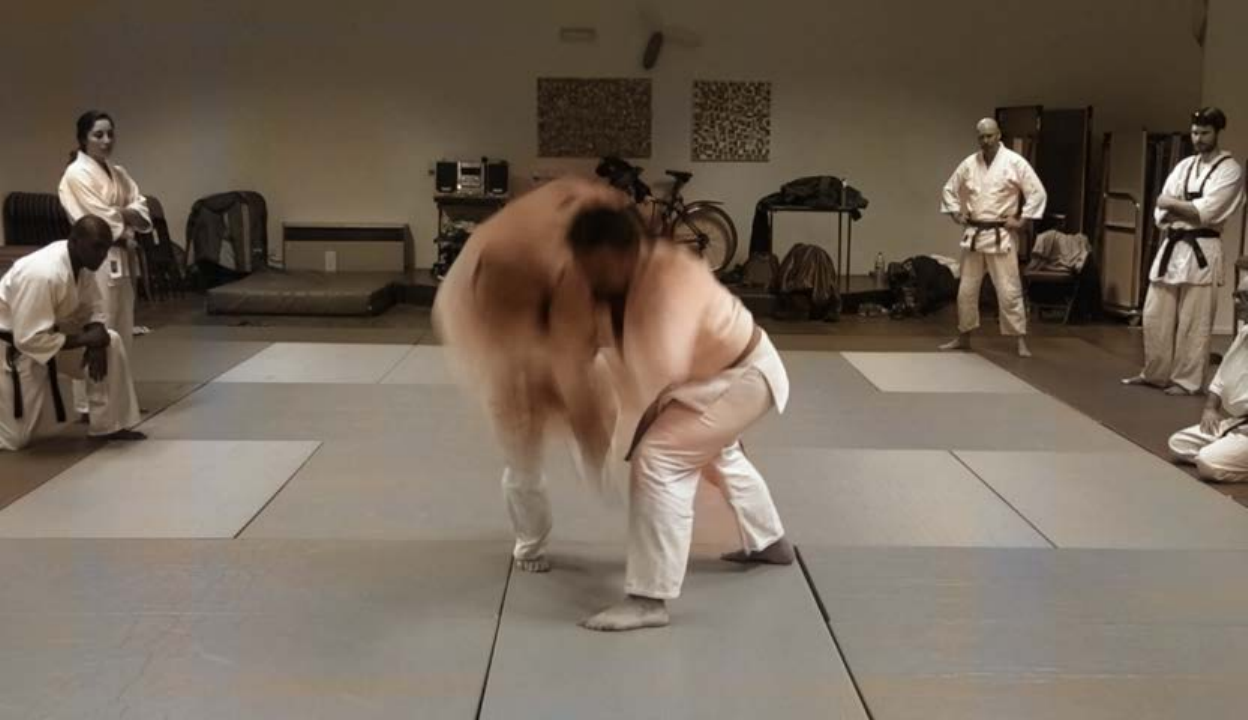}
		&\includegraphics[width=0.115\textwidth]{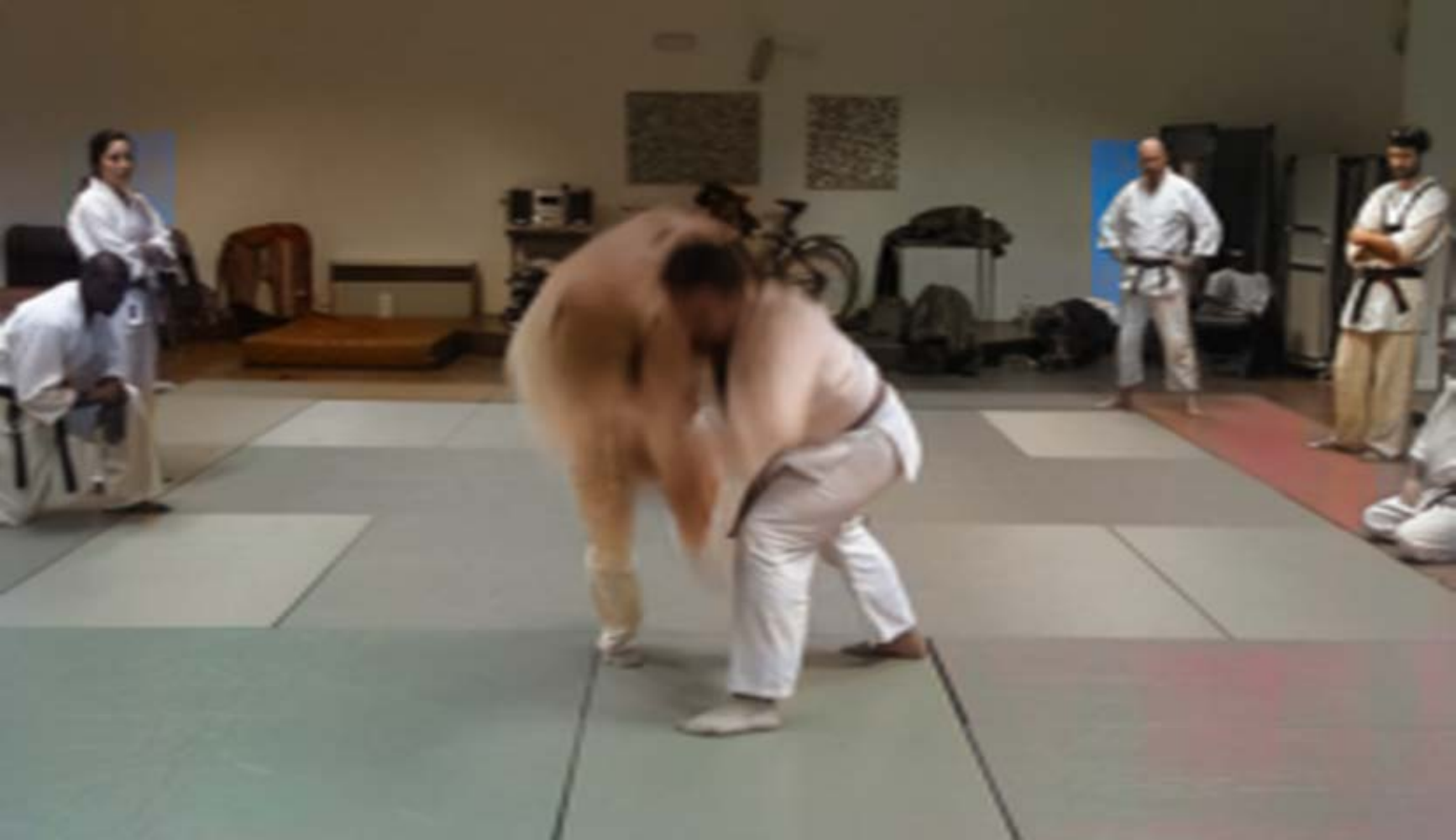}
		&\includegraphics[width=0.115\textwidth]{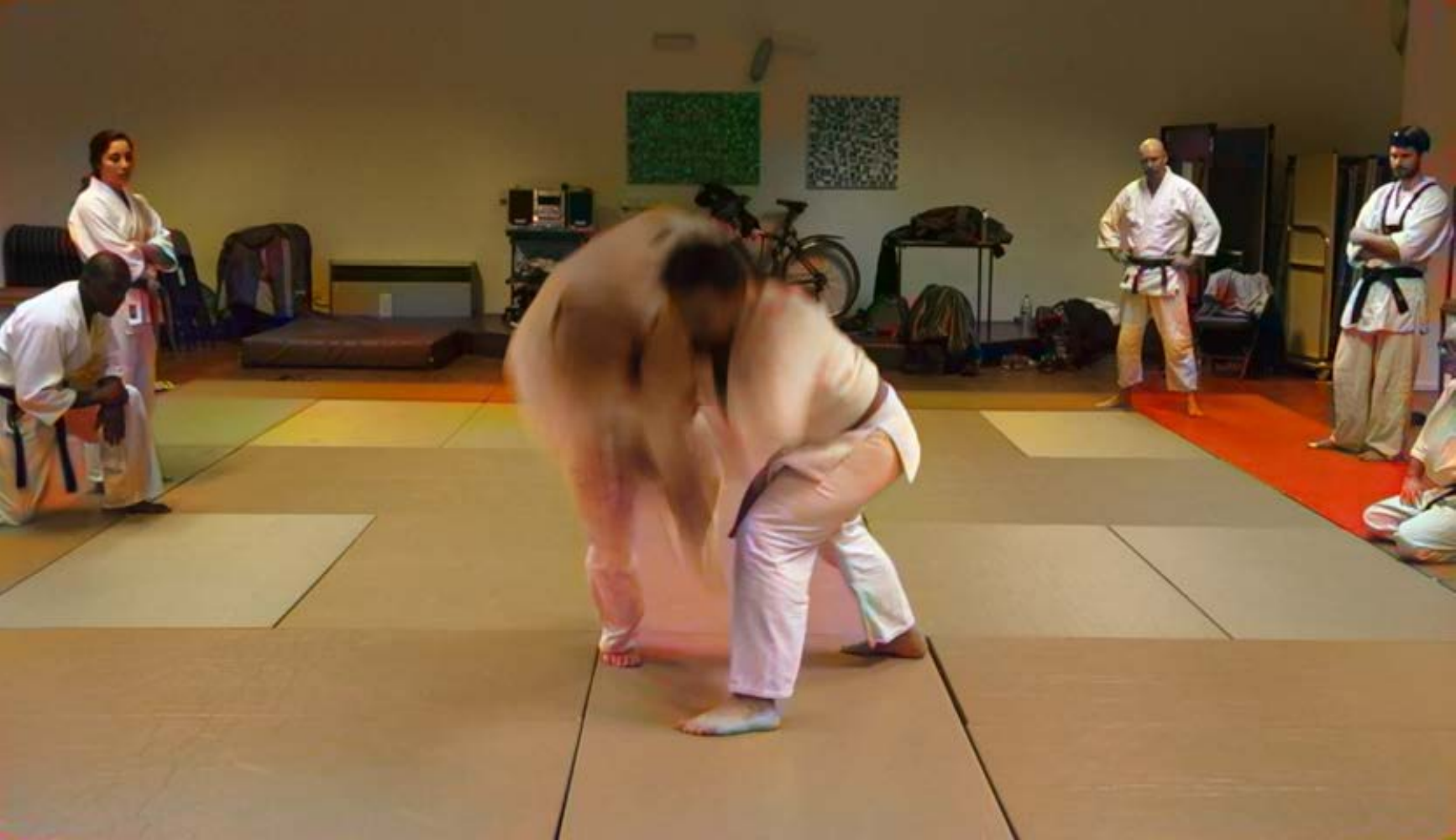}
		&\includegraphics[width=0.115\textwidth]{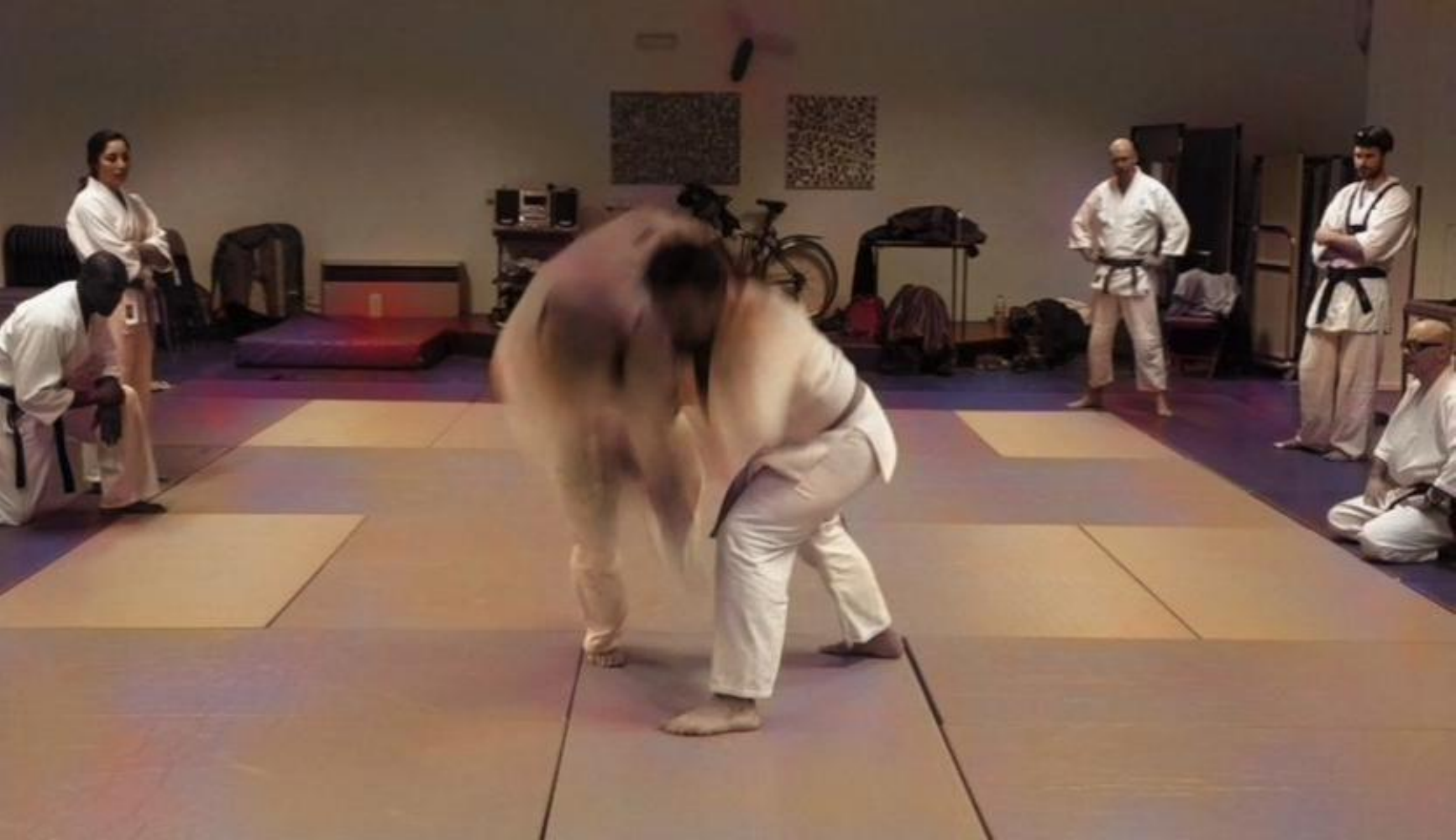}
		&\includegraphics[width=0.115\textwidth]{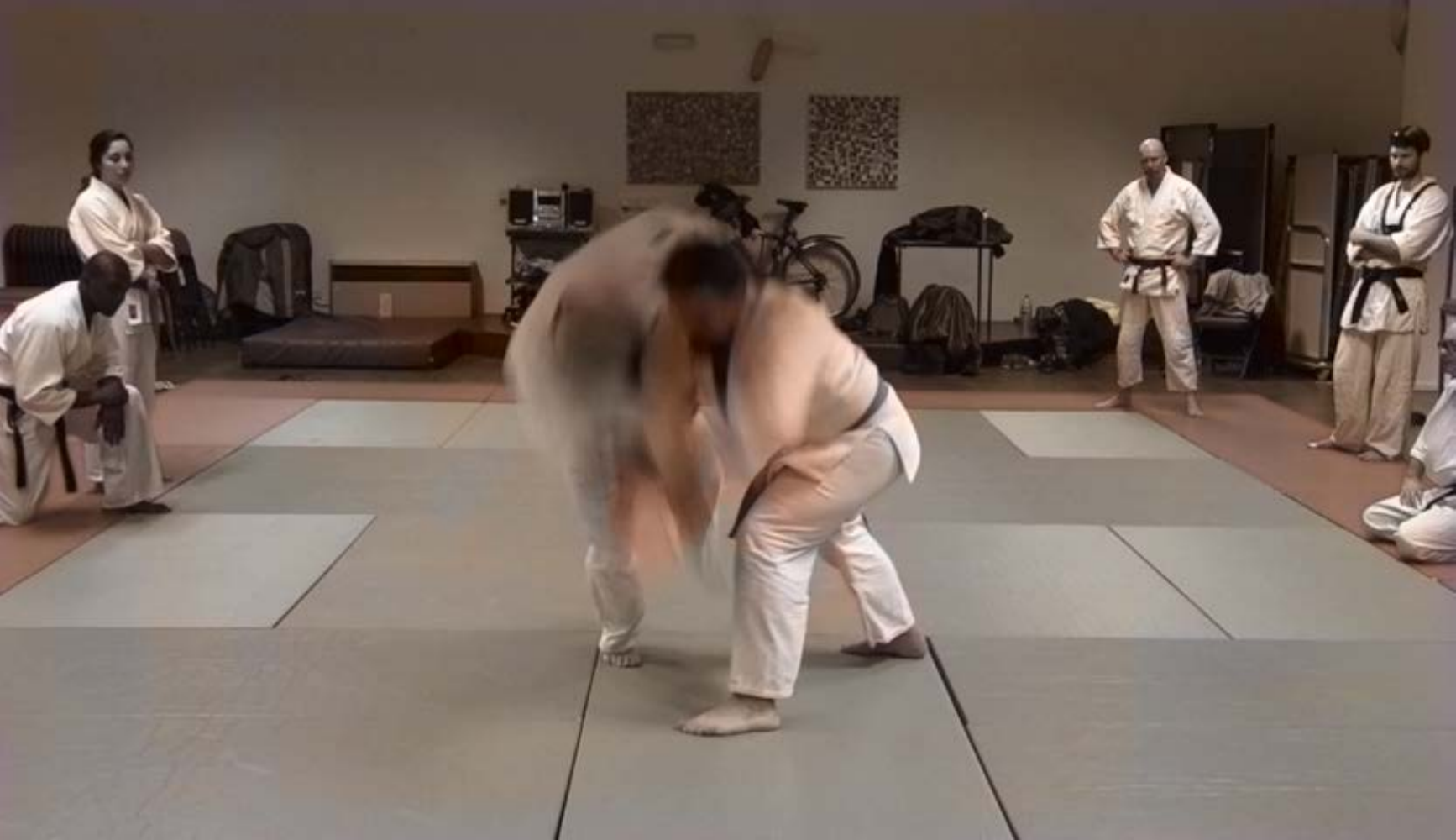}
		&\includegraphics[width=0.115\textwidth]{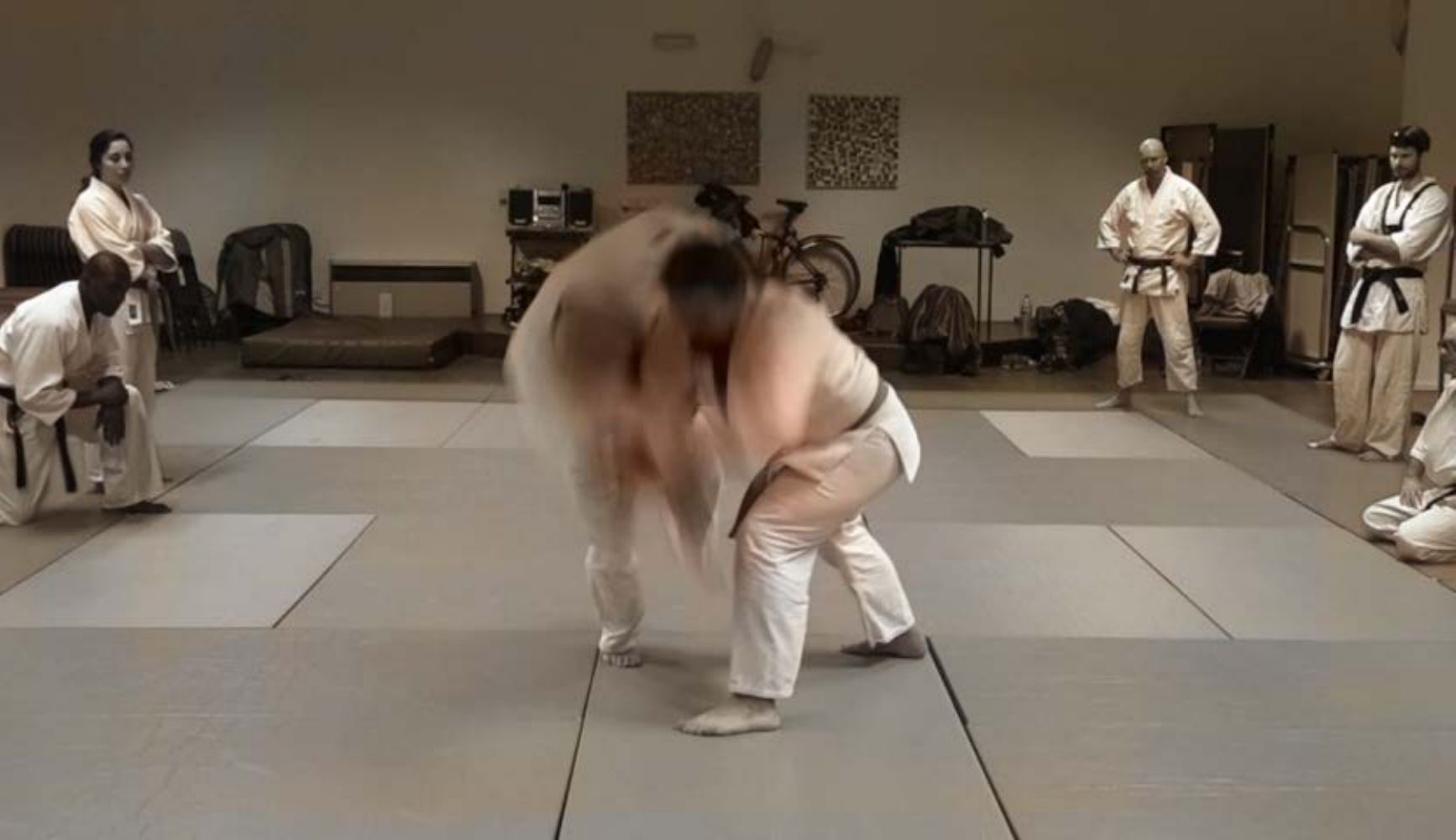}
		&\includegraphics[width=0.115\textwidth]{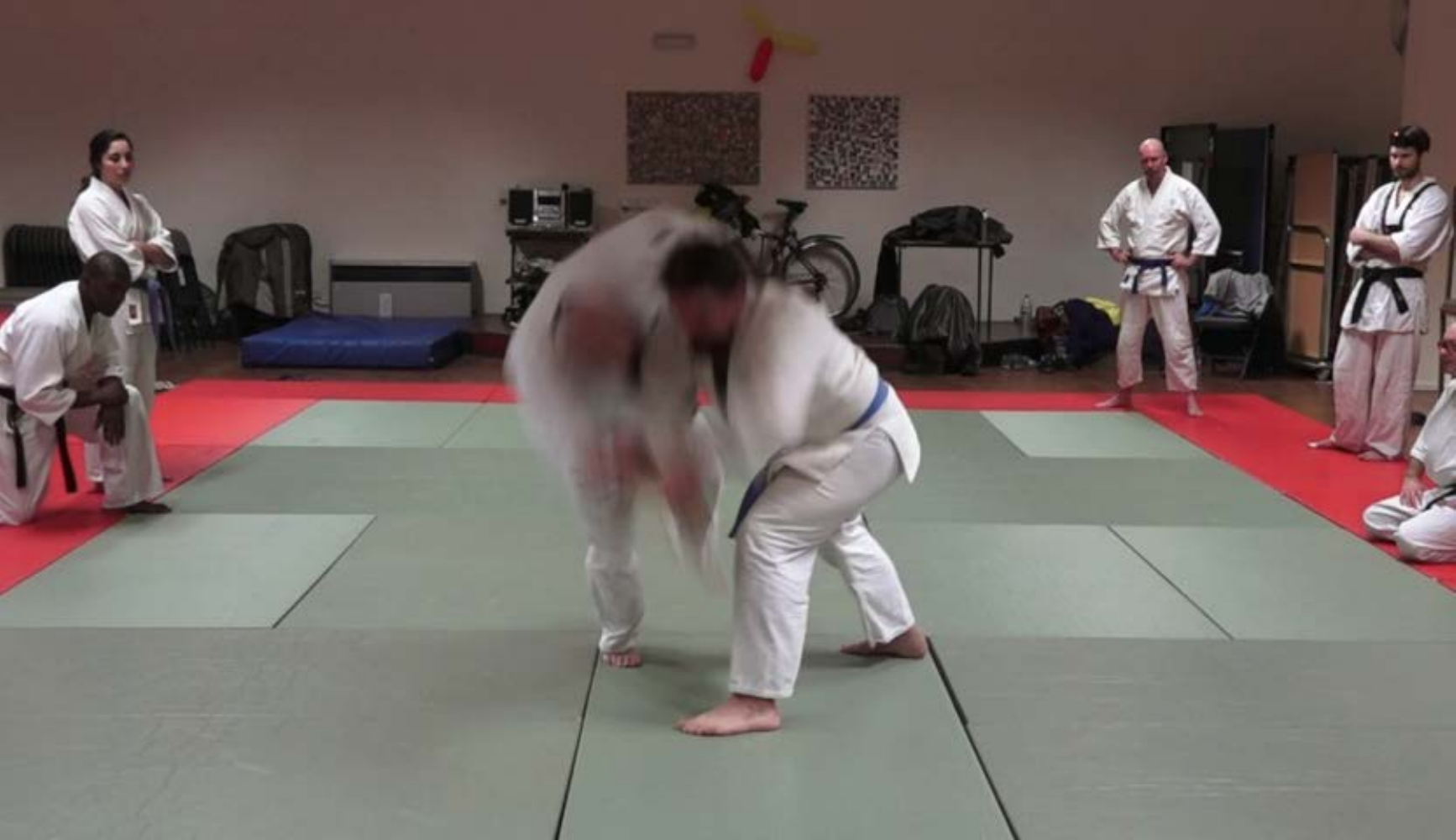}
		&\includegraphics[width=0.115\textwidth]{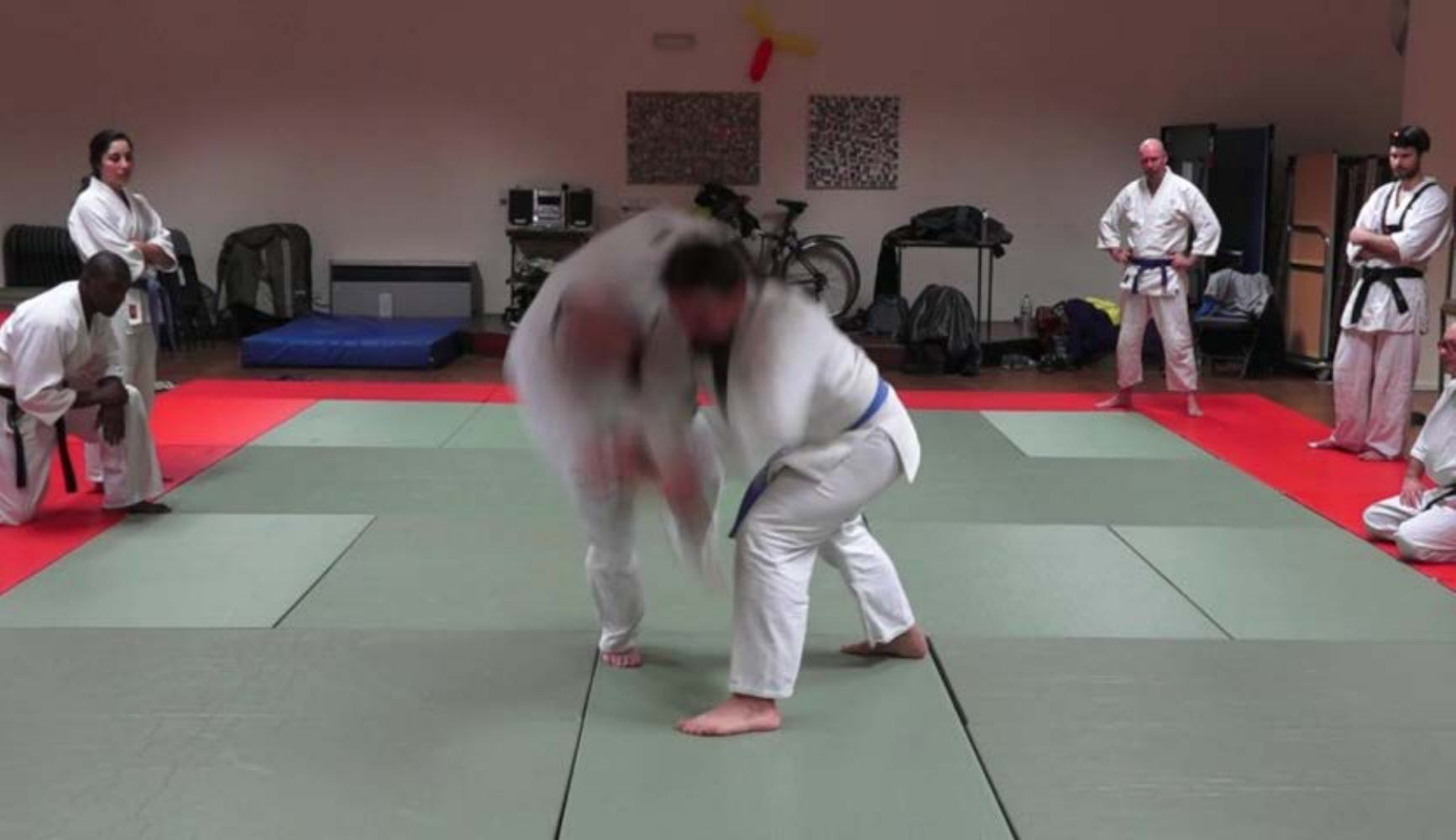}
		\\
		&\includegraphics[width=0.115\textwidth]{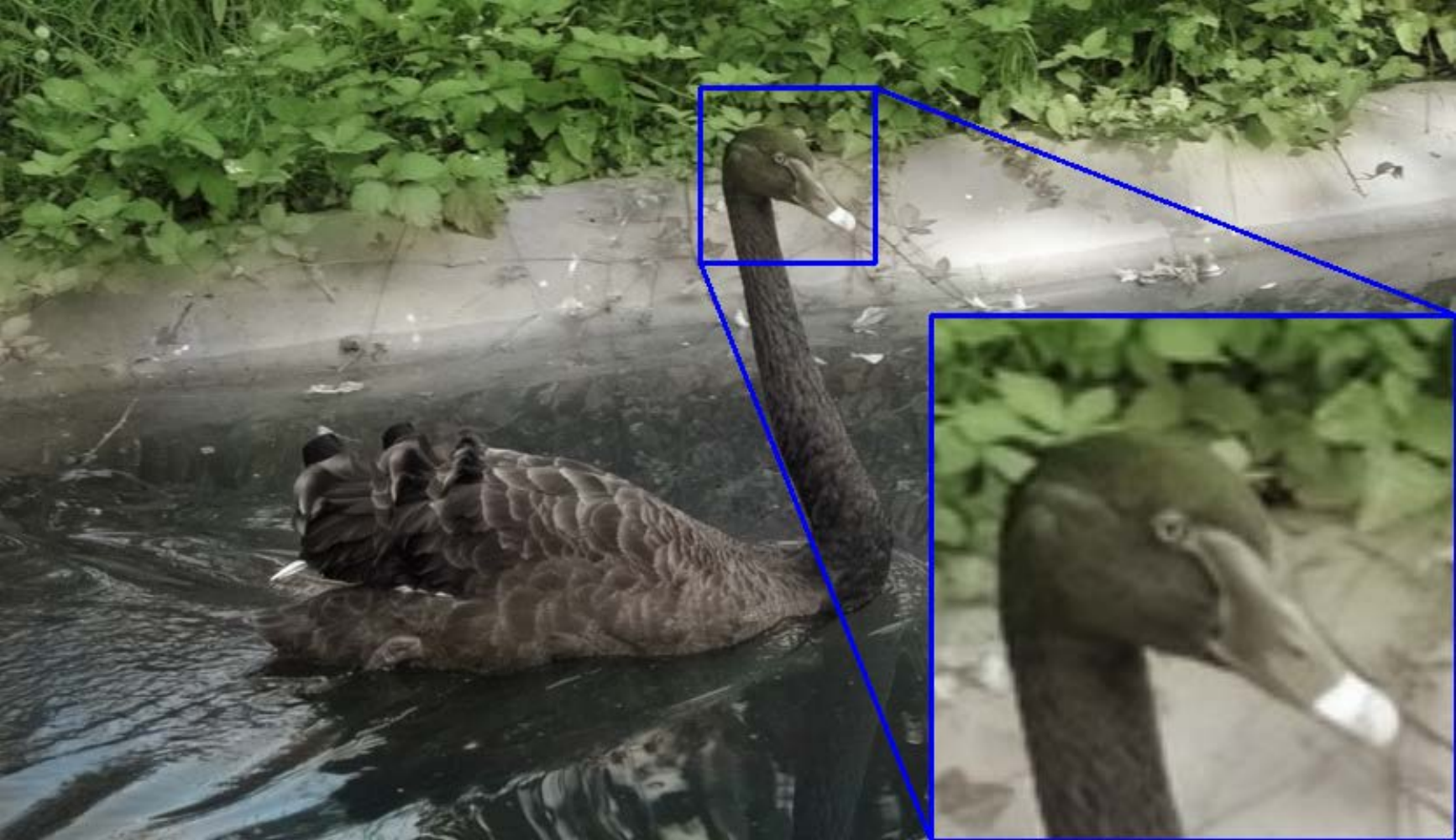}
		&\includegraphics[width=0.115\textwidth]{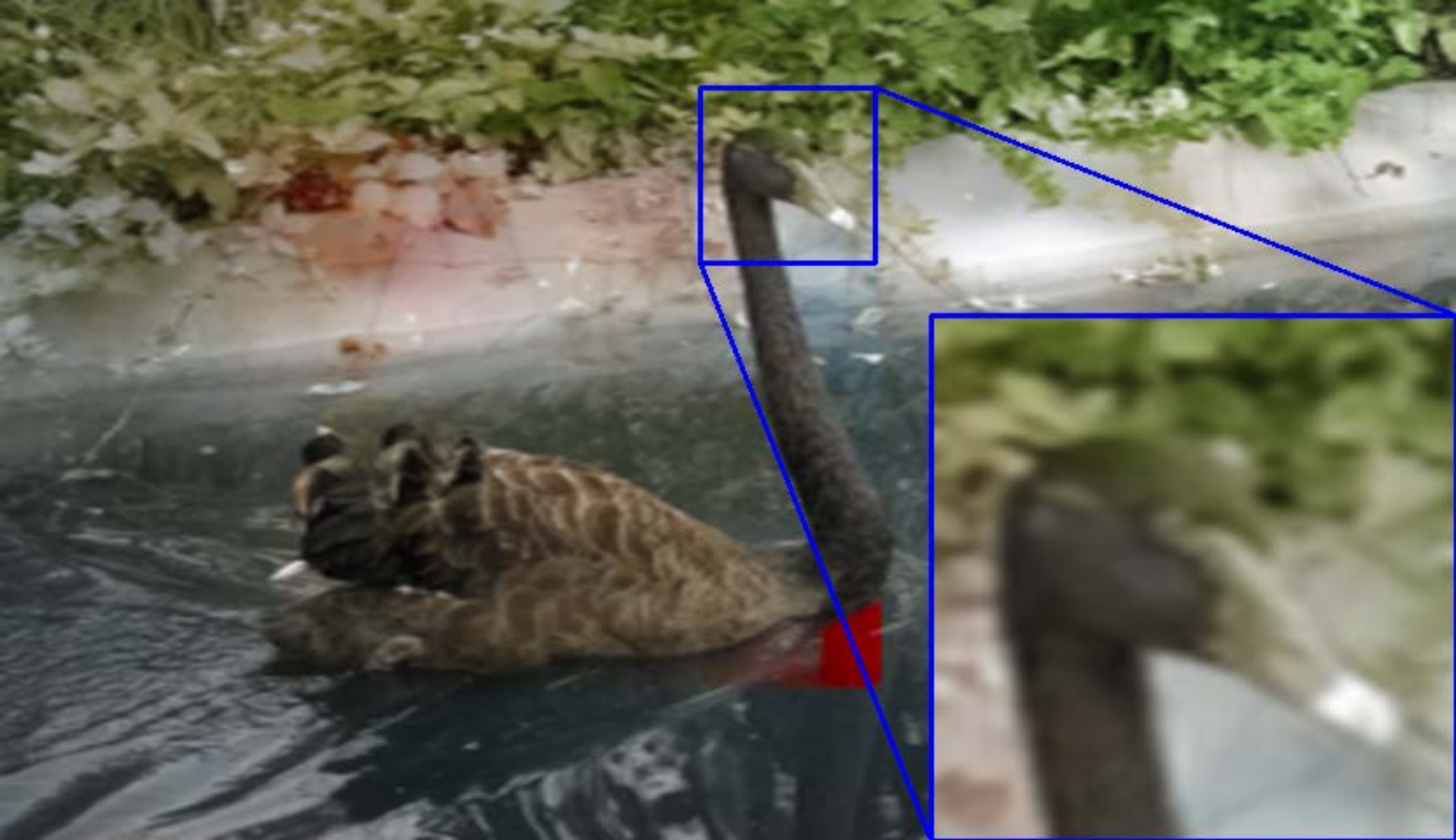}
		&\includegraphics[width=0.115\textwidth]{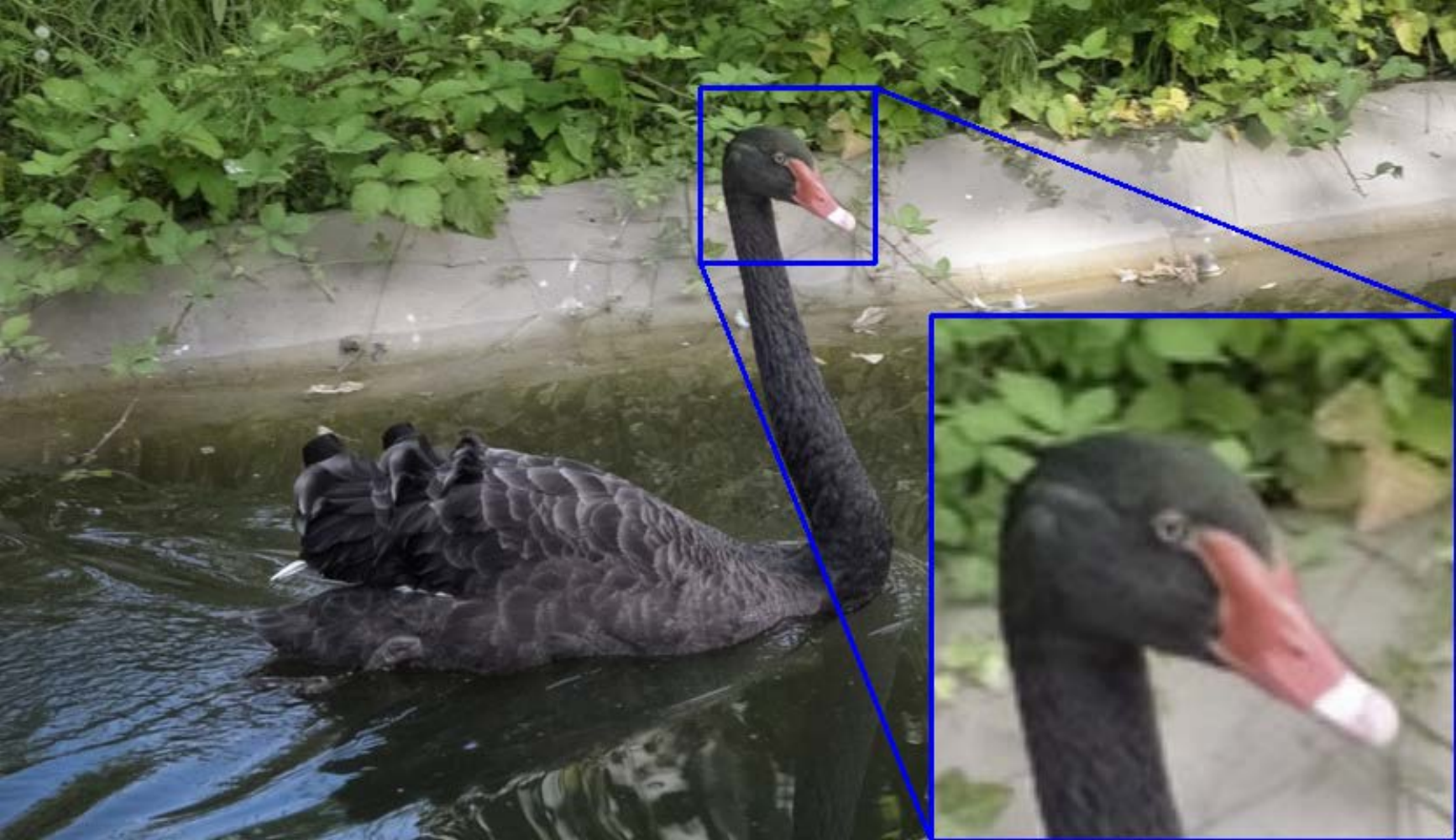}
		&\includegraphics[width=0.115\textwidth]{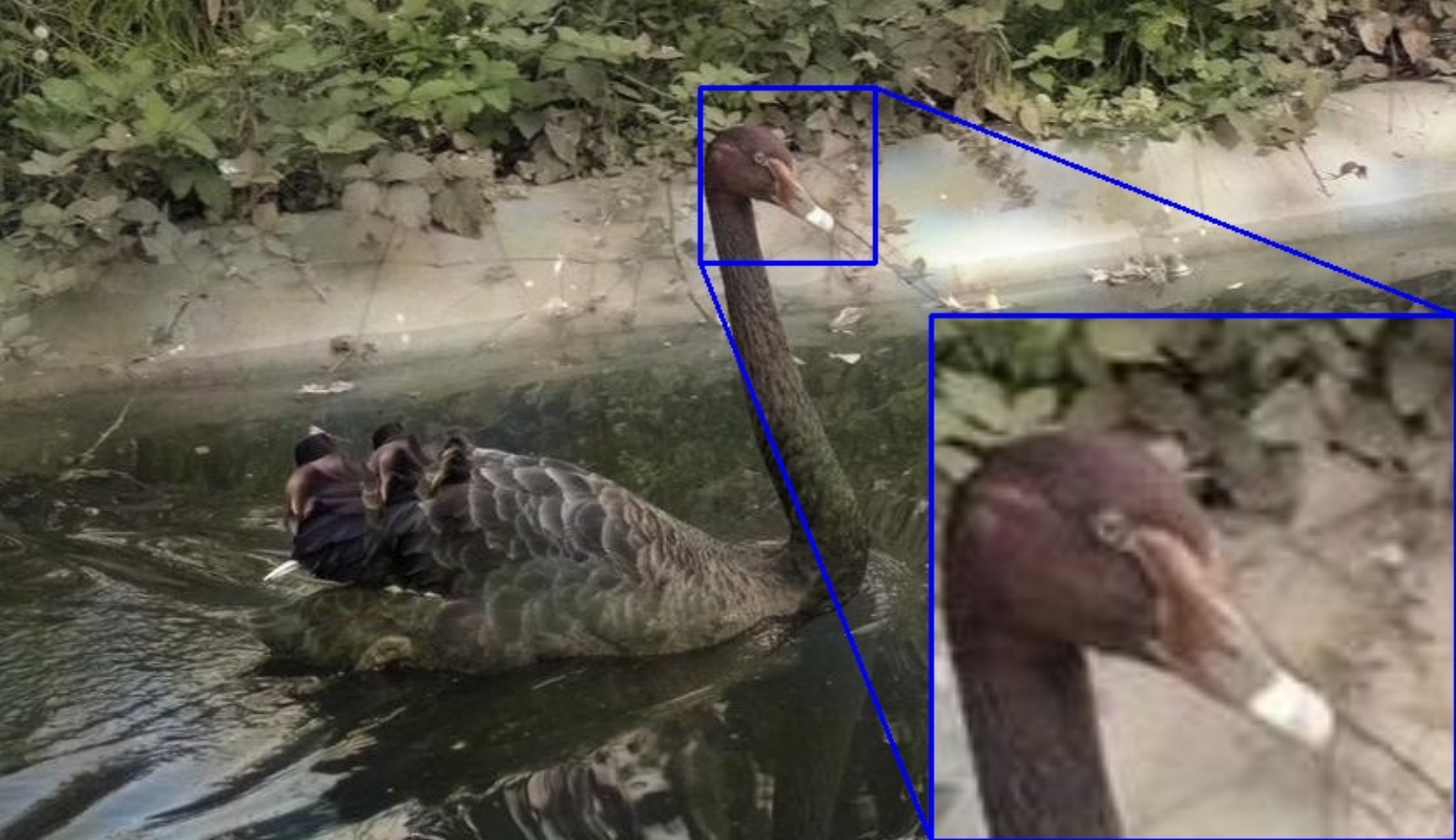}
		&\includegraphics[width=0.115\textwidth]{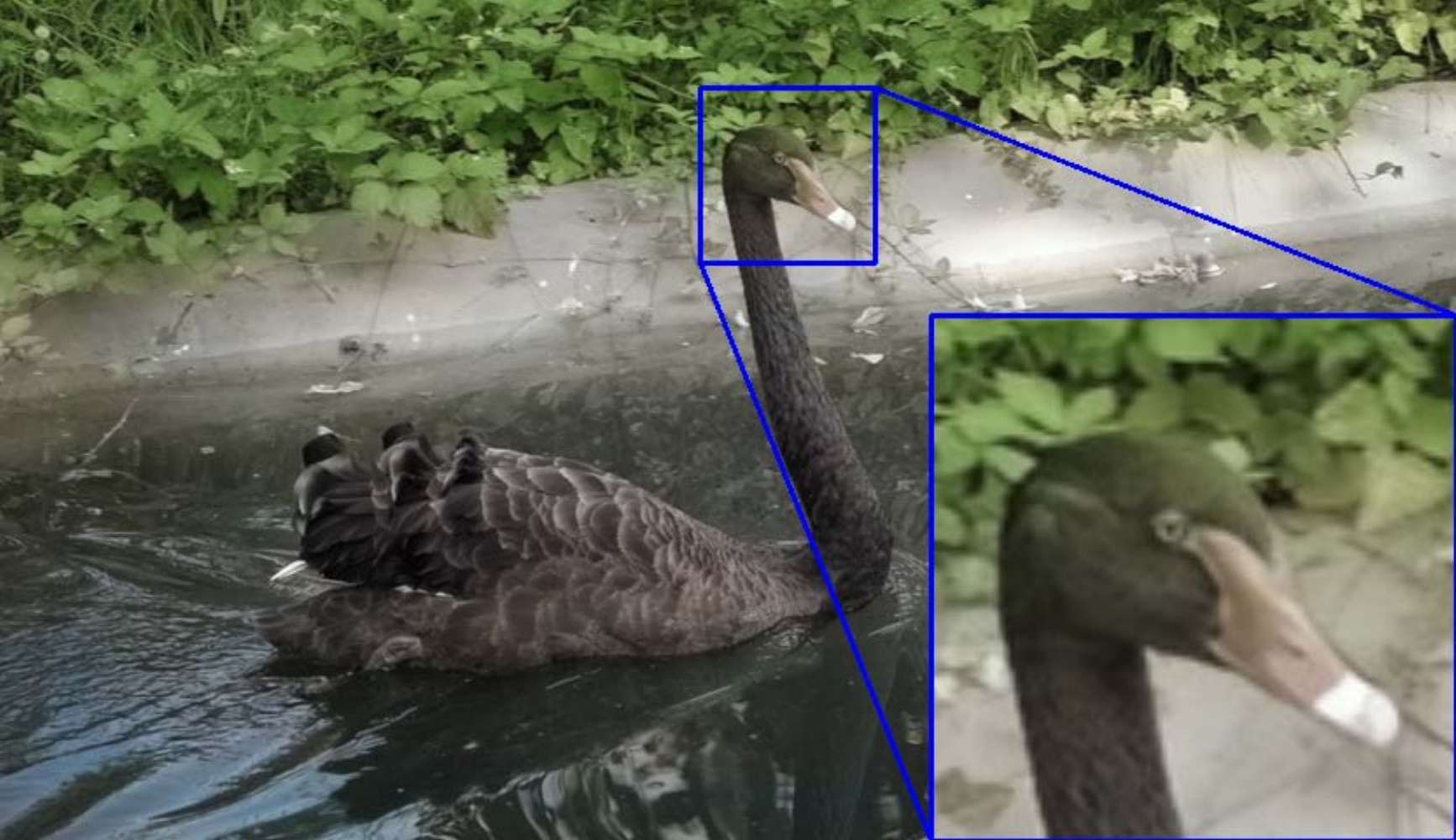}
		&\includegraphics[width=0.115\textwidth]{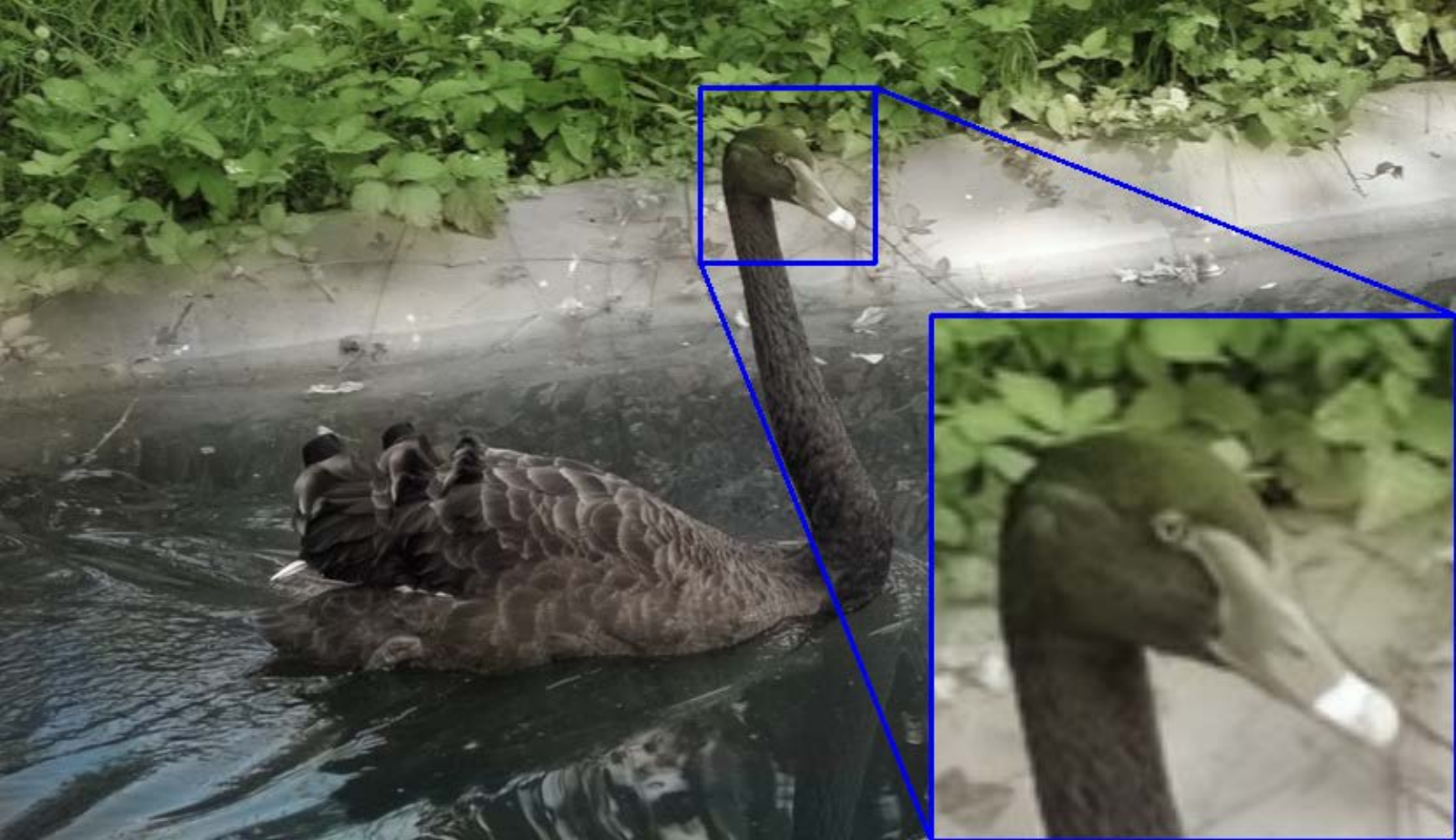}
		&\includegraphics[width=0.115\textwidth]{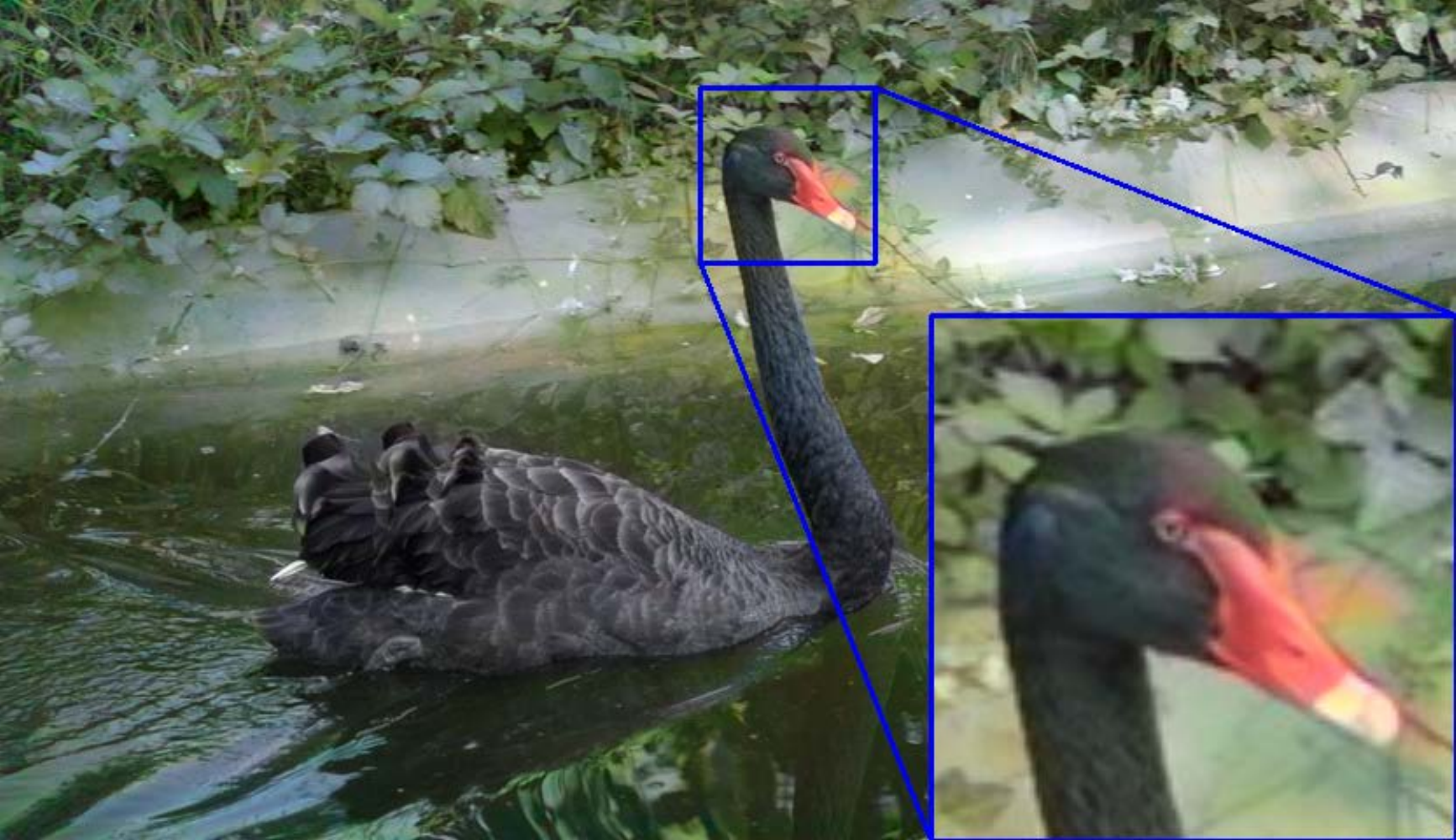}
		&\includegraphics[width=0.115\textwidth]{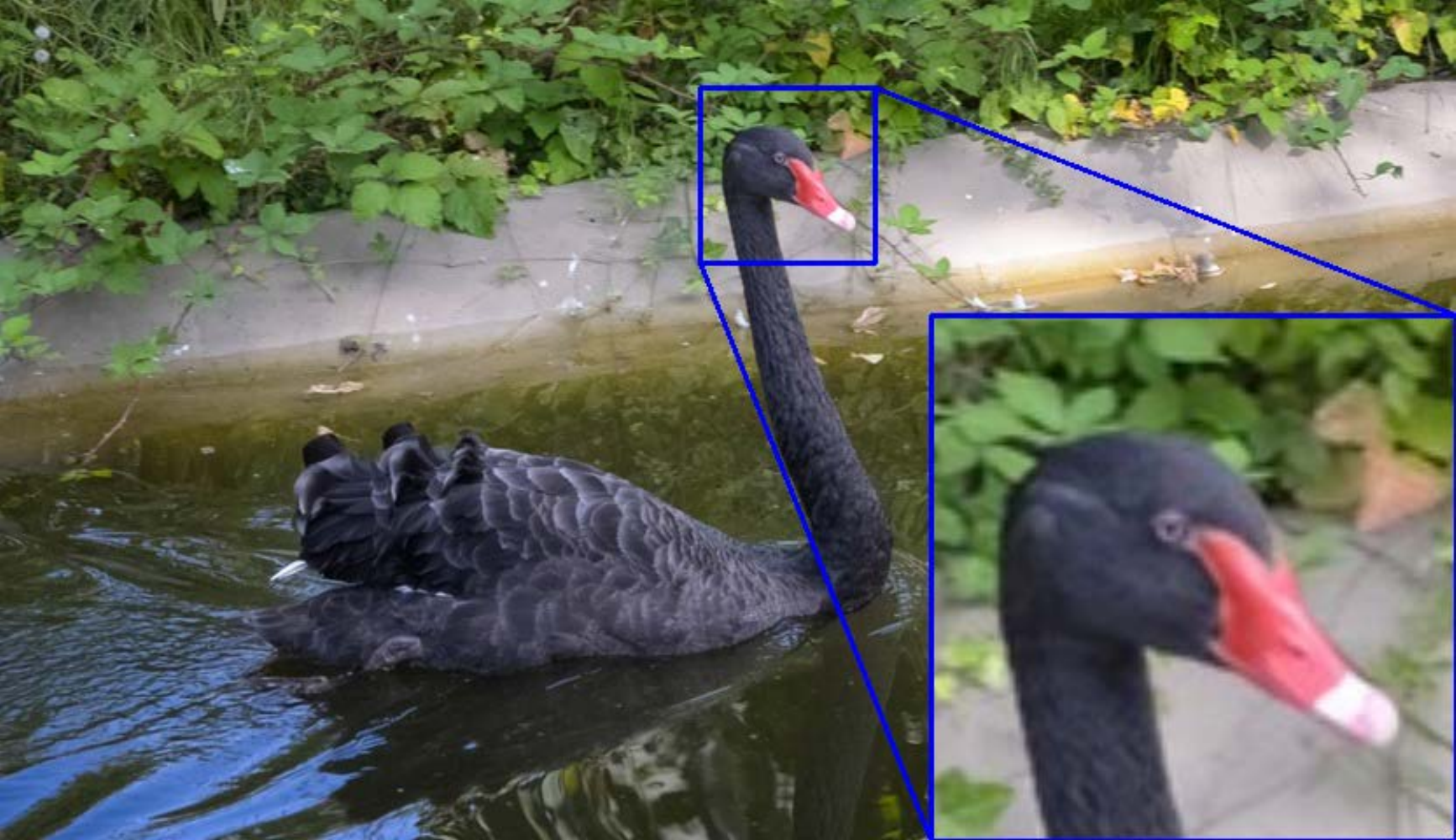}
		\\
		&\footnotesize IDC &\footnotesize InstColor &\footnotesize GCP & \footnotesize IDC+BTC &\footnotesize FAVC & \footnotesize TCVC & \footnotesize  ST-HVC (Ours)& \footnotesize GT\\ 
	\end{tabular}
	\caption{Chroma comparison with representing and state-of-the-art approaches. From top to bottom, the figure demonstrates outstanding performance of the proposed method regarding the front color, background color, blurred pixels and the detail.}
	\label{fig:compare_color} 
\end{figure*}

\subsection{Joint Flow and Histogram Feature Module}{\label{jfhm}}
The designed joint flow and histogram feature module (JFHM) is tailored to utilize the chroma and motion feature condensed in histogram and flow as depicted in Fig.~\ref{fig:stahg}, frames features are first put into the module. Flow and histogram information is integrated in SA and feature refinement. 

Inspired by \cite{tran2015learning,qiu2017learning,zhang2022flow,shao2022decoupled}, we decouple the attention schema along the spatial and temporal dimension, thus forming the temporal attention (TA) and spatial attention (SA).
Specifically, given the input feature $F^i, i\in \left [t-\tau, t+\tau \right ]$, we split it into $s^2$ non-overlapped windows along the height and width dimension where each window of $F^i$ is denoted as $F^i_{jk}$, $j, k \in \left [1, s\right ]$. We take different window size for each attention and $s_{temporal}$ is smaller than $s_{spatial}$. 

The temporal approach group windows along temporal dimension, i.e., $G_{temporal} = \{F^{t-\tau}_{j, k},..., F^{t+\tau}_{j, k}\}, j, k \in \left [1, s_{temporal}\right ]$ so that temporal attention is performed across tokens in $G_{temporal}$. Thus, continuous movement of the object inside a small spatial window can be detected. 

In spatial branch, we gather them along spatial dimension $G_{spatial} = \{F^{i}_{1, 1},F^{i}_{1, 2},..., F^{i}_{s_{spatial}, s_{spatial}}\}, i\in \left [t-\tau, t+\tau \right ]$, and attention is performed across tokens in it. The operation helps our model to understand similar textures in spatially-neighboring pixels.


Giving a sequence of frame features
$\phi_x$, the output after concatenating the TA and SA features $M_1$ can be described:
\begin{equation}
	M_1 = \phi_x \oplus TA(LN(\phi_x) \oplus SA(LN(\phi_x),F_{flow})
\end{equation}
where LN means the Layer Normalization and $F_{flow}$ stands for flow features of the frames. The sign $\oplus$ denotes the concatenate operation.
After concatenation, we integrate the histogram feature and we introduce the skip connection after the feature refinement and feed forward layer. 

The histogram reference serves as a multi-scale color distribution indicator. Following~\cite{chen2007real,wang2022local} and by splicing the pixel histogram along the range dimension, we can obtain different histogram patch and attain distinct feature map at multiple scale.
The HistConv in Fig.~\ref{fig:overview} denotes the $2 \times 2$ convolution without bias and with its weight shrunk to downscale the hist-map.
The output of joint feature module $M_2$ is:
\begin{equation}
	M_2 = M_1 \oplus FFN(LN(M_1 \oplus M_{FR}))
\end{equation}
where $FFN$ stands for feed forward network. $M_{FR}$ means the middle output after integrating histogram and feature refinement. 
The structure of feature refinement module is composed of three times duplicated sequence of $3 \times 3$ convolution and LeakyReLU, where we inject color information of reference histogram into grayscale frame features.

\subsection{Feature Recombination via U-shape Network}
The concatenation results from Sec.\ref{input} are put into the designed U-shape network which consists of basic encoder-decoder, JFHM module and a histogram exploitation module. 

Motivated by ~\cite{zhang2022histogram,afifi2021histogan,wang2022local}, we design to exploit the reference histogram from the middle frame of video. The JFHM aims to integrate the flow and histogram information, and help the reconstruction process.
As is shown in Fig.~\ref{fig:overview}, we apply JFHM in skip connection and replace the bottleneck with it as well.  
Besides, we employ dynamic region-aware convolution~\cite{chen2021dynamic} in our decoder to learn different kernels according to local illumination features. The DRBlock in the legend of Fig.~\ref{fig:overview} is composed of a sequence of original DR convolution, batch normalization and ReLU, repeated twice.

\begin{table*}[htb!]
	\centering
	\caption{Quantitative comparisons on DAVIS and Videvo datasets. We pick both image and video-based approaches for comparisons. The best result is highlighted in black bold while the second best is marked underline.}
	\label{tab:quantitative}
	\resizebox{\linewidth}{!}{
		\begin{tabular}{c | cccc | cccc}
			\hline	
			\toprule	
			\multirow{2}{*}{Methods} & \multicolumn{4}{c|}{DAVIS} & \multicolumn{4}{c}{Videvo} \\ 
			& PSNR $\uparrow$ &  SSIM $\uparrow$ & Warp Error $\downarrow$ & L2 Error $\downarrow$ & PSNR $\uparrow$ & SSIM $\uparrow$ &Warp Error$\downarrow$ & L2 Error $\downarrow$  \\ 
			\hline
			\midrule
			CIC~\cite{zhang2016colorful} 	  & 22.77 & 0.9431 & 0.06055 & 15.88 & 22.56 & 0.9417 & 0.03317 & 17.11 \\ 
			IDC~\cite{zhang2017real} 	  & 24.85 & 0.9436 & 0.05377 & 11.99 & 25.17 & 0.9568 & 0.02997 & 11.69 \\ 
			InstColor~\cite{su2020instance} & 24.51 & 0.9411 & 0.07828 & 13.20 & 24.80 & 0.9458 & 0.04917 & 20.37 \\ 
			GCP~\cite{wu2021towards} 	  & 23.53 & 0.9309 & 0.04978 & 12.45 & 24.25 & 0.9369 & 0.02864 & 12.08 \\ 
			\midrule
			CIC~\cite{zhang2016colorful}+BTC~\cite{lai2018learning}   & 22.11 & 0.9298 & 0.05170 & 16.67 & 21.77 & 0.9343 & 0.02891 & 17.68 \\ 
			IDC~\cite{zhang2017real}+BTC~\cite{lai2018learning}   & 23.91 & 0.9006 & {\bf 0.04498} & 12.91 & 23.74 & 0.9383 & \underline{0.02786} & 12.60\\ 
			FAVC~\cite{lei2019fully}	  & 22.98 & 0.9055 & 0.06002 & 13.26 & 23.47 & 0.9183 & 0.03236 & 12.21 \\ 
			TCVC~\cite{liu2021temporally}	  & 25.49 & 0.9550 & 0.04819 & 11.86 & \underline{25.43} & 0.9570 & 0.03589 & 11.59 \\ 
			\midrule
			Ours	  & {\bf 26.68} & {\bf 0.9612} & \underline{0.04626} &{\bf  10.38} & {\bf 26.95} & {\bf 0.9623} & {\bf 0.02612} & {\bf 10.13} \\ 
			Ours w/o histogram	  & \underline{25.62} & \underline{0.9590} & 0.05052 &\underline{11.49} & 25.22 & \underline{ 0.9587} & 0.02963 & \underline{ 11.37} \\ 
			\bottomrule
		\end{tabular}
	}
\end{table*}

\subsection{Loss Functions}
Inspired by ~\cite{lai2018learning,liu2021temporally}, our model aims to reduce the temporal flickers via adopting the warping loss $L_{w}$ as:
\begin{equation}
	L_{w}=\sum_{d=\{1,2\}} \sum_{t=1}^{N-d}\left\|M_{t+d \Rightarrow t} \odot\left(y_{t}-y_{t+d}^{\text {warp }}\right)\right\|_{2}
\end{equation}
where $d$ stands for the frame interval, thus the loss under larger interval means longer temporal coefficient. $y_t$ means the $t_{th}$ frame and $M_{t+d \Rightarrow t}=\exp \left(-\alpha\left\|y_{t}-y_{t+d}^{\text {warp }}\right\|_{2}^{2}\right)$ represents the visibility mask~\cite{lai2018learning}.  $y_{t+d}^{\text {warp }}=\operatorname{\mathcal{W}}\left(y_{t+d}, f_{t+d \rightarrow t}\right)$ where $\mathcal{W}$ warps the $(t+d)_{th}$ frame under the indication of flow $f_{t+d \Rightarrow t}$ from $(t+d)_{th}$ frame to $t_{th}$ frame. 

Besides, we adopt the Charbonnier loss $L_c = \sum_{t=1}^{N} \sqrt{(\hat{y_t}-y_t)+\epsilon^2}$ and smooth loss $L_s$ in $Lab$ space to avoid the gradient exploding and produce smooth result. The overall loss function is shown below:
\begin{equation}
	L_{total}=\lambda_1 L_w + \lambda_2 L_c + L_s
\end{equation}
where $\lambda_1$ and $\lambda_2$ denote the weights of loss functions.

\section{Experiment}
\subsection{Experimental Procedure}
{\bf Dataset and Metrics.} 
Following previous works ~\cite{lai2018learning, lei2019fully,zhao2022vcgan}, we adopt DAVIS dataset ~\cite{perazzi2016benchmark} and Videvo dataset ~\cite{lai2018learning} for training and testing.
Originally designed for video segmentation, DAVIS dataset includes a variety of moving objects and distinct motion types. It contains 60 short videos for training and 30 for testing, with 100 frames in each video. The Videvo dataset has 80 long videos for training and 20 for testing. There are about 300 frames in each video clips.
To evaluate the quality of generated video frames, we use PSNR, SSIM, warp error~\cite{lai2018learning} and L2 error. The warp error is to measure the temporal continuity of the generated frames.

{\bf Implementation Details. }
We utilize the pytorch framework to implement our model with Adam~\cite{kingma2014adam} its optimizer. A single GeForce RTX 3090 graphic card is used to train and test the models. During the training, the size of input frames are resized to 256 * 256. The learning rate is initialized to 5e-5 and we set the $\alpha$ in loss functions to 9 according to ~\cite{lai2018learning}.

\begin{figure}[h]
	\centering
	\begin{tabular}{c}
		\includegraphics[width=0.45\textwidth]{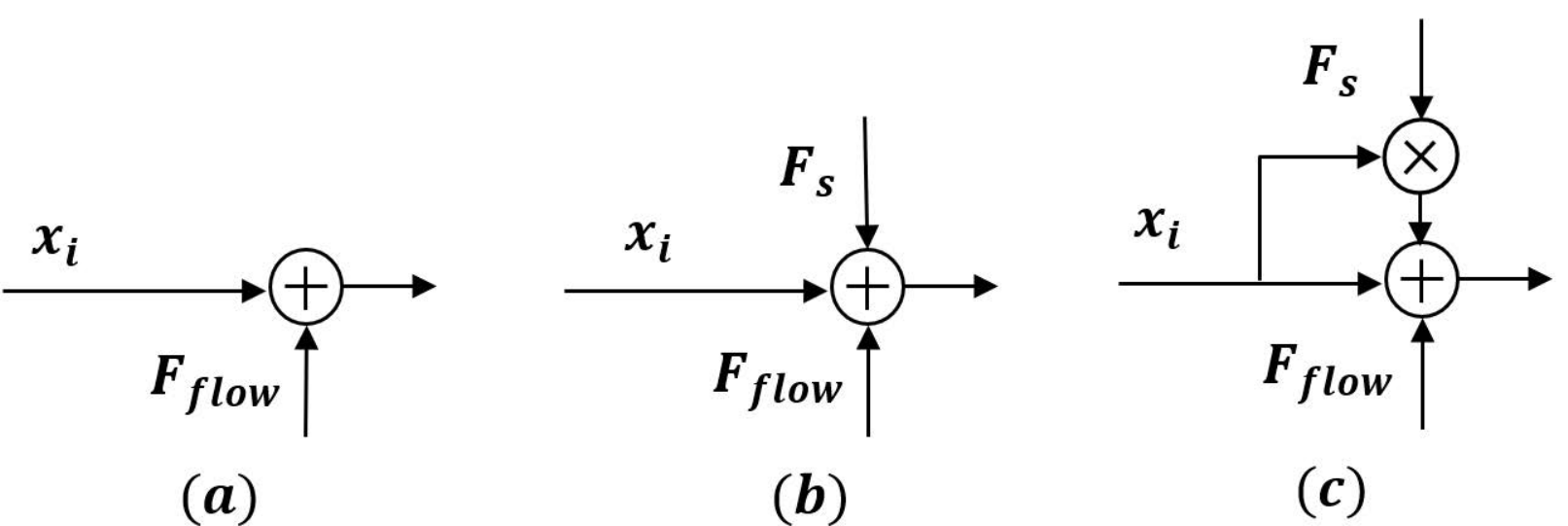}
	\end{tabular}
	\caption{Illustrations of each feature connection schema where (c) is employed in our proposed ST-HVC.}
	\label{fig:aba_input}
\end{figure}

\begin{table}[h]
	\centering
	\caption{Ablation study on feature connection schema. Detailed structures of (a), (b) and (c) are defined in Fig~\ref{fig:aba_input}.}
	\begin{tabular}{c|c c c}
		\toprule
		Settings & (a) & (b) & (c)\\ 
		\hline 
		\midrule
		PSNR	      & 26.12  & 26.49 & 26.95 \\ 
		drop rate 	  & 3.08\%  & 1.71\% & - \\
		\bottomrule
	\end{tabular}
	\label{tab:aba_input}
\end{table}

\begin{table}[!ht]
	\centering
	\caption{Ablation study on key components. Experiments are conducted in Videvo dataset.}
	\begin{tabular}{c|c c c}
		\hline
		\toprule
		Settings  & PSNR$\uparrow$ & SSIM$\uparrow$ & Warp Error $\downarrow$\\ 
		\hline
		\midrule
		w/o histogram guide		  & 25.22  & 0.9590 & 0.02993 \\ 
		w/o flow integration  & 26.31  & 0.9577 & 0.03836 \\
		w/o spatial attention & 24.39  & 0.9370 & 0.03295 \\
		w/o temporal attention & 26.16  & 0.9551 & 0.04122 \\
		ours  	   	  & 26.95  & 0.9623 & 0.02612 \\
		\bottomrule
	\end{tabular}
\label{tab:aba_components}
\end{table}

\begin{figure}[t]
\centering
\begin{tabular}{c}
	\includegraphics[width=0.45\textwidth]{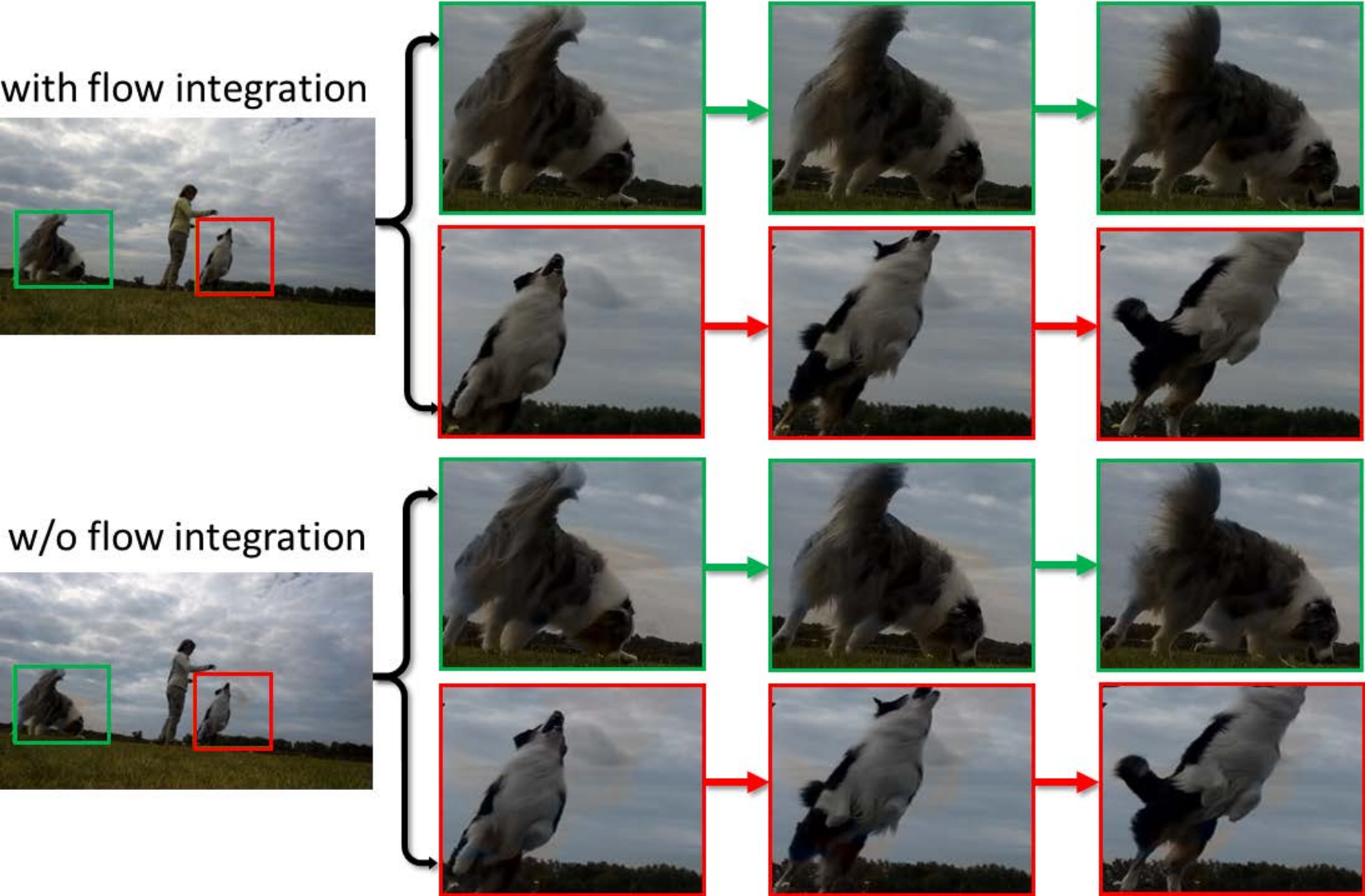}
\end{tabular}
\caption{Visual results of ablation on flow integration. $\rightarrow$ denotes the movement of frames. Caused by inherent attributes of optical flow, artifacts occur without flow integration. }
\label{fig:aba_flow_integration}
\end{figure}

\begin{figure}[t]
\centering 
\begin{tabular}{c@{\extracolsep{0.2em}}c@{\extracolsep{0.2em}}c@{\extracolsep{0.2em}}c@{\extracolsep{0.2em}}c}
	
	\rotatebox{90}{\footnotesize InstColor}
	
	&\includegraphics[width=0.105\textwidth]{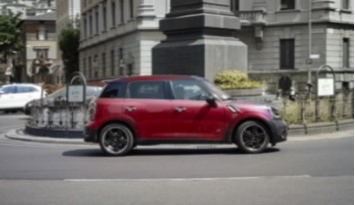}
	&\includegraphics[width=0.105\textwidth]{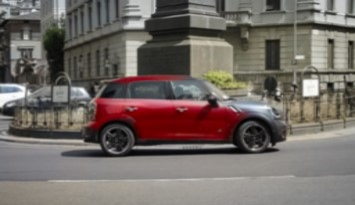}
	&\includegraphics[width=0.105\textwidth]{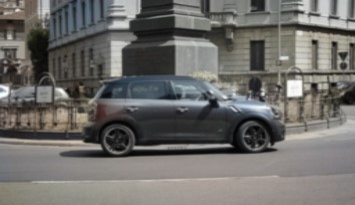}
	&\includegraphics[width=0.105\textwidth]{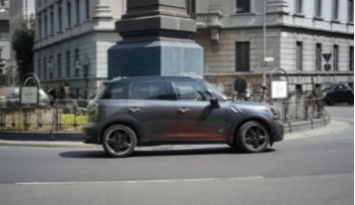}
	\rotatebox{90}{\;\footnotesize 1.91e-1}
	\\
	\rotatebox{90}{\;\;\footnotesize GCP}
	
	&\includegraphics[width=0.105\textwidth]{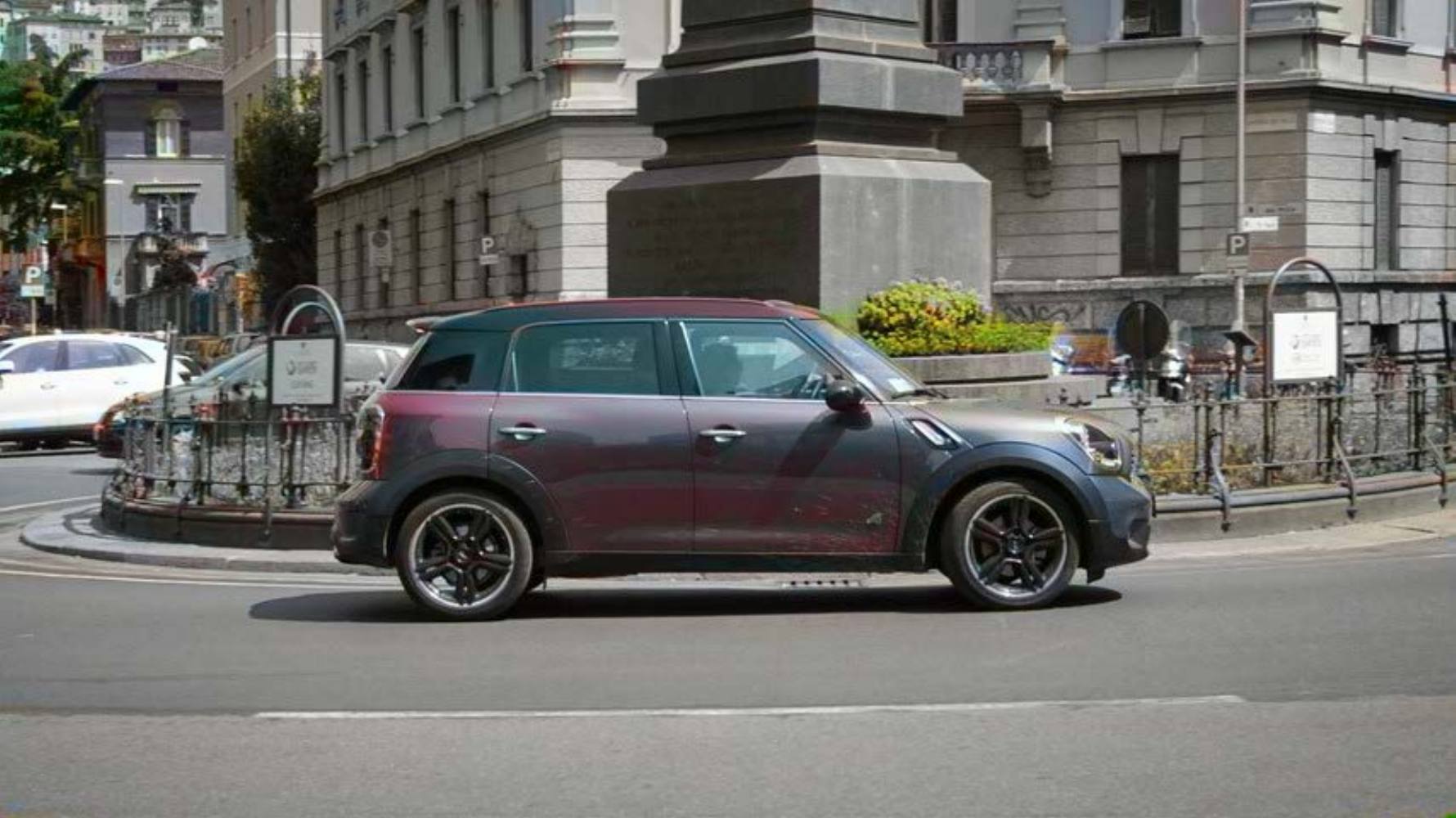}
	&\includegraphics[width=0.105\textwidth]{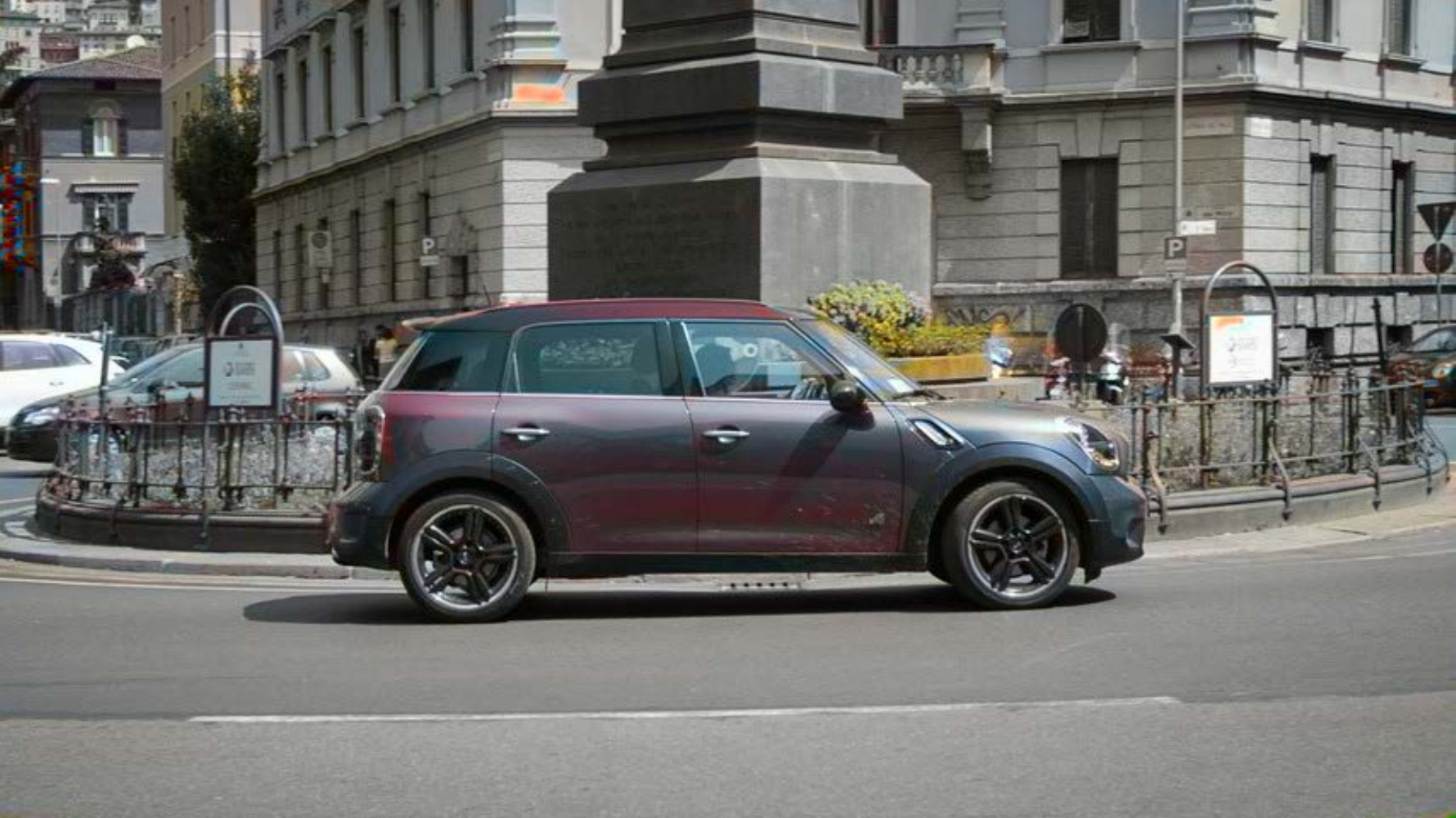}
	&\includegraphics[width=0.105\textwidth]{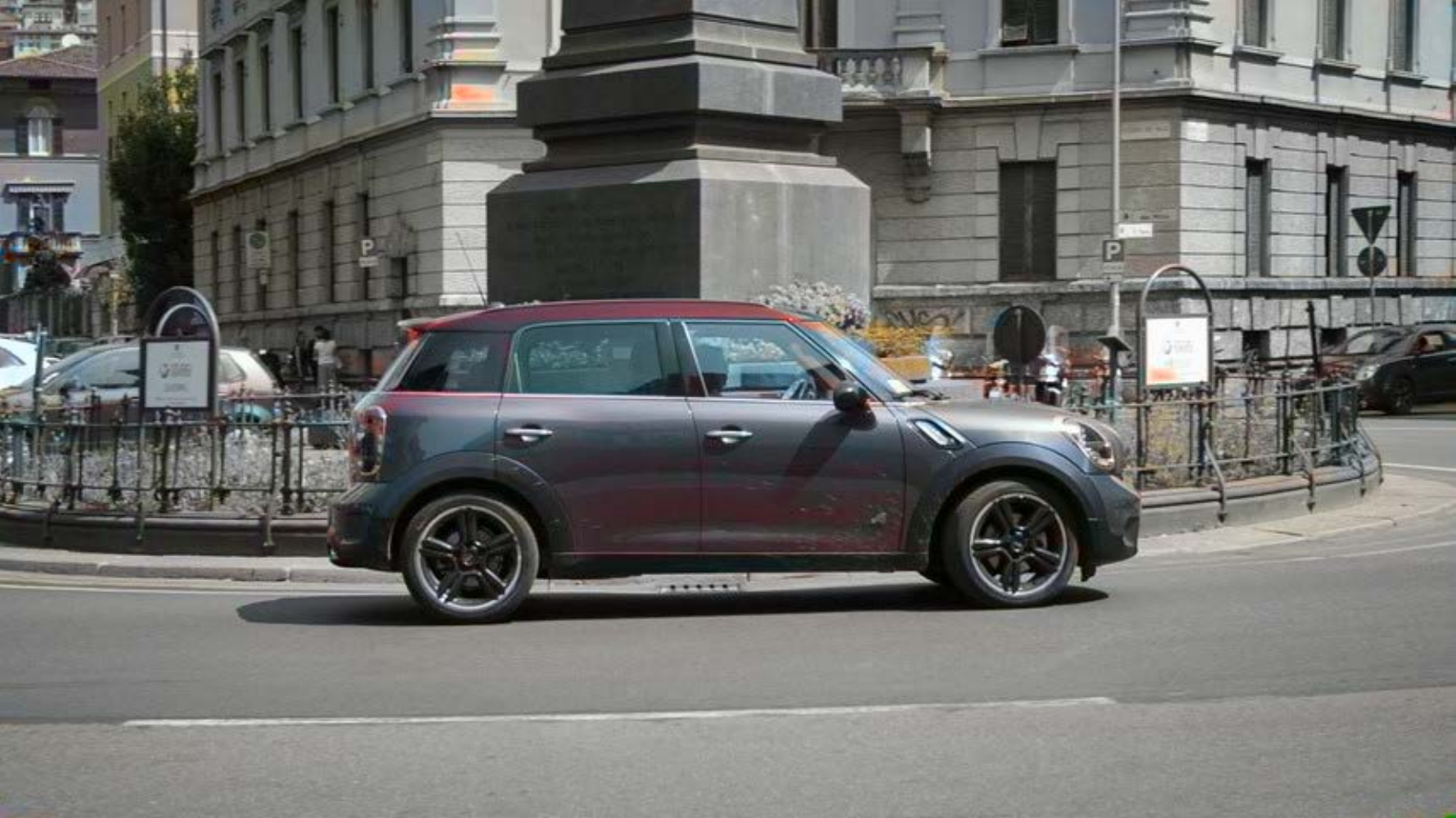}
	&\includegraphics[width=0.105\textwidth]{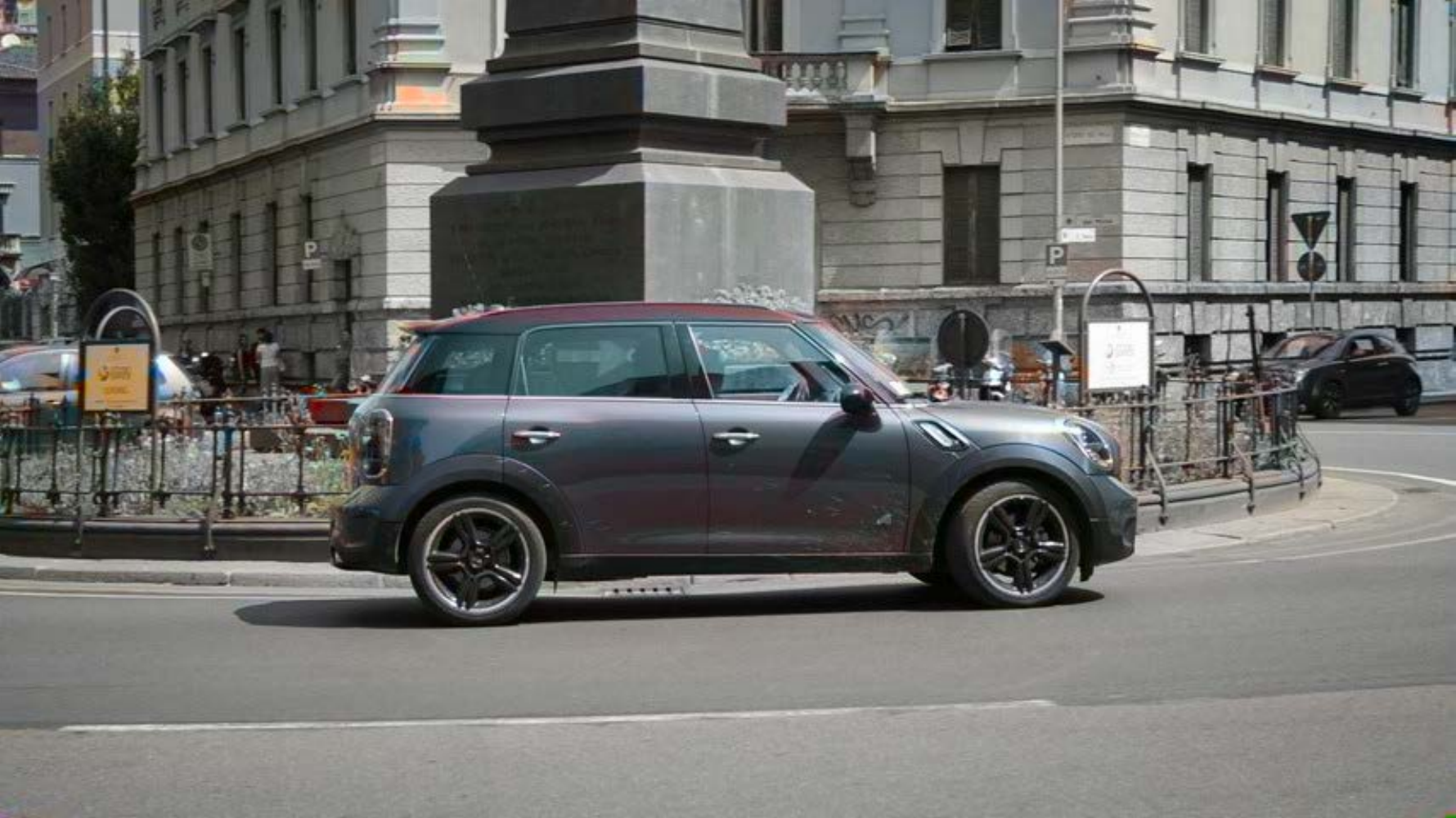}
	\rotatebox{90}{\;\footnotesize 1.79e-1}
	\\
	
	\rotatebox{90}{\;\footnotesize TCVC}
	
	&\includegraphics[width=0.105\textwidth]{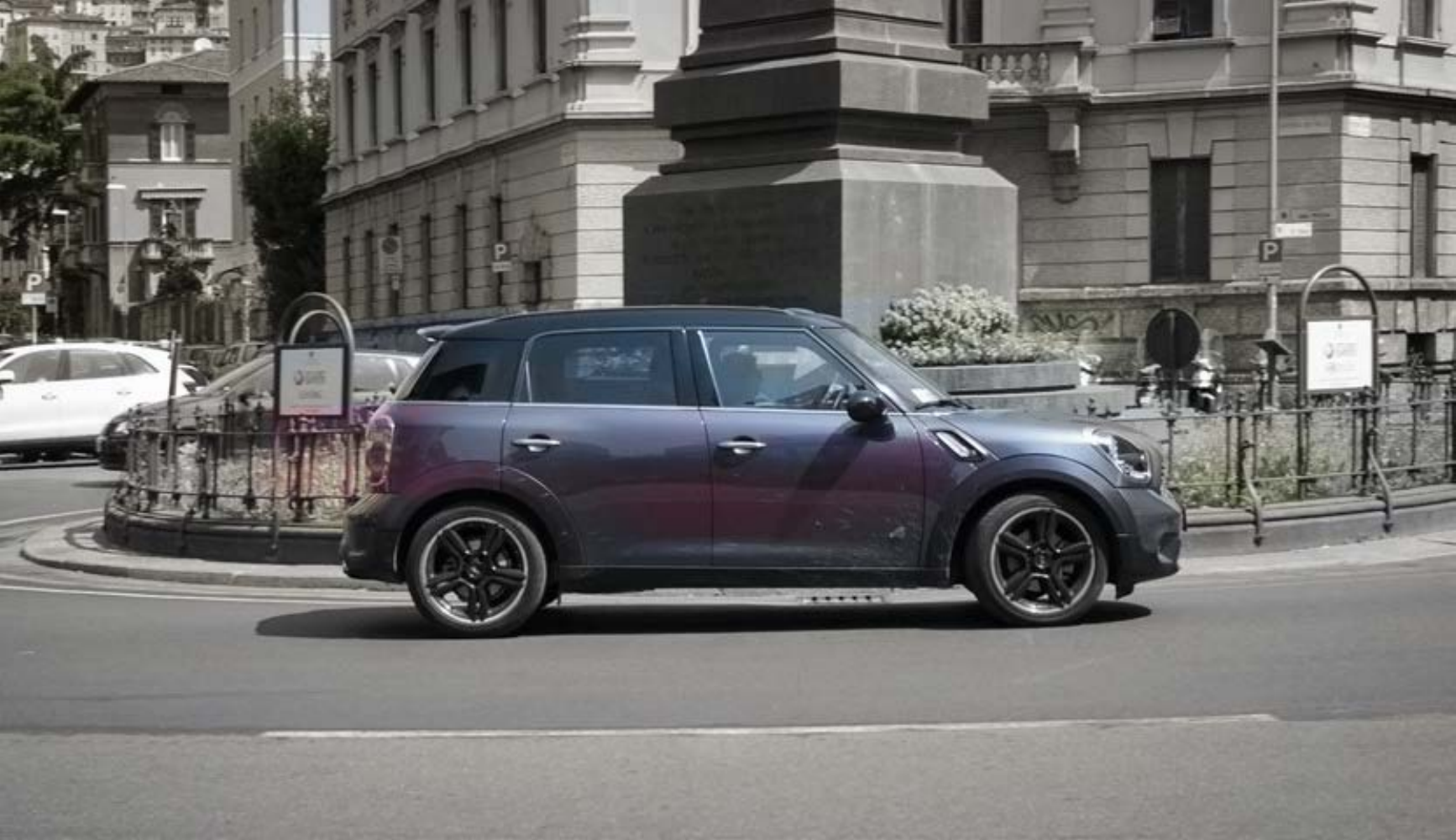}
	&\includegraphics[width=0.105\textwidth]{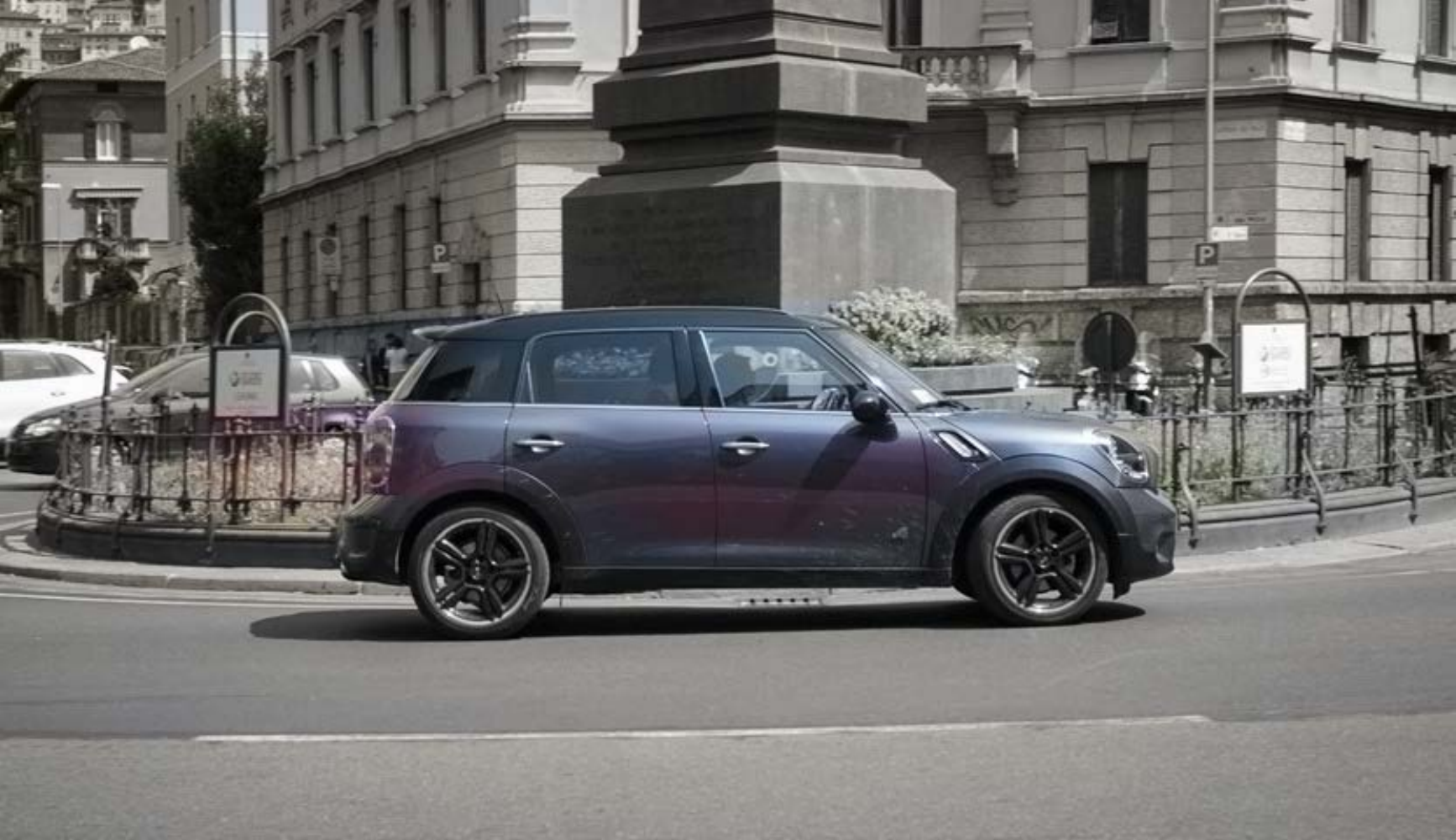}
	&\includegraphics[width=0.105\textwidth]{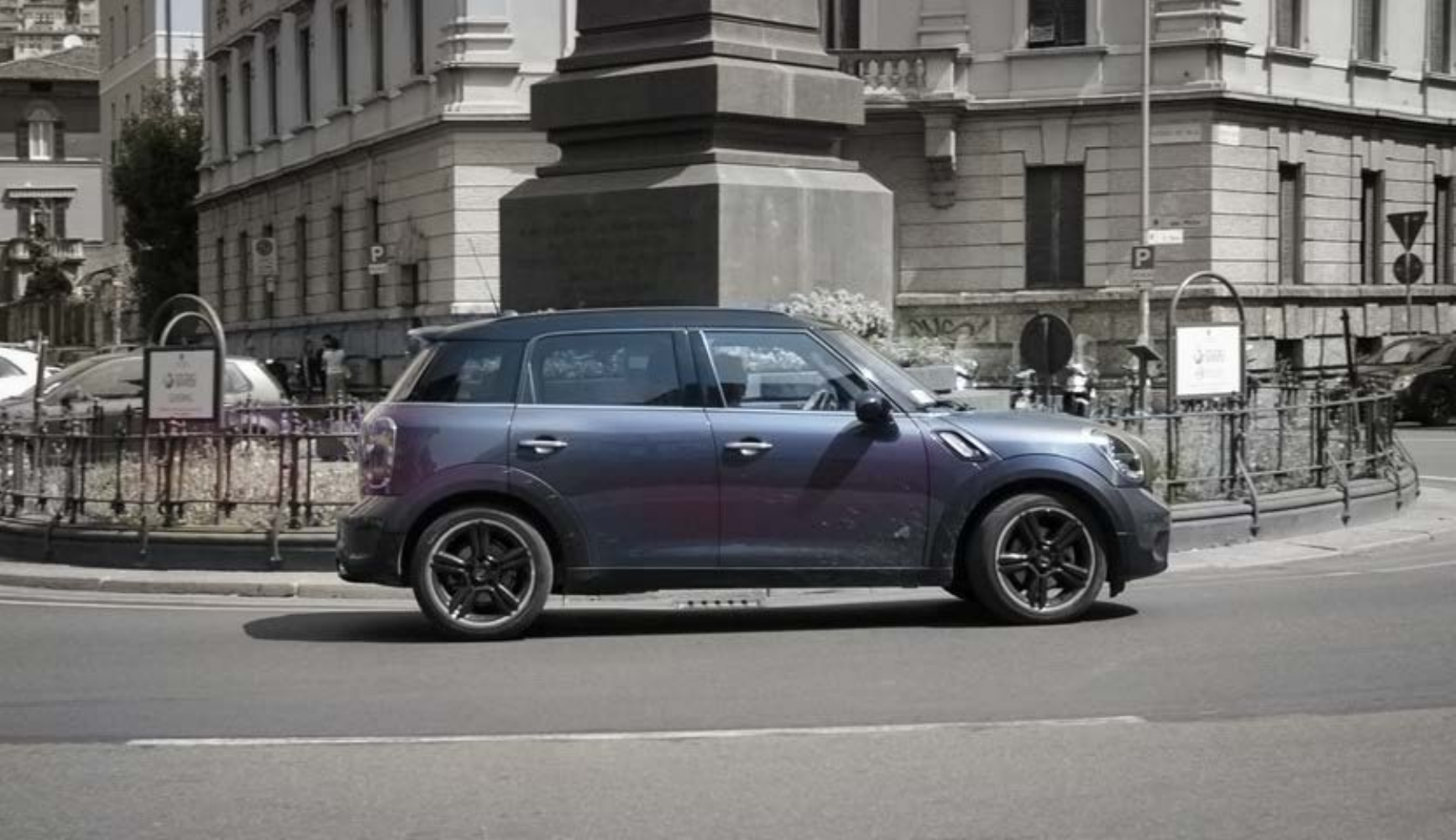}
	&\includegraphics[width=0.105\textwidth]{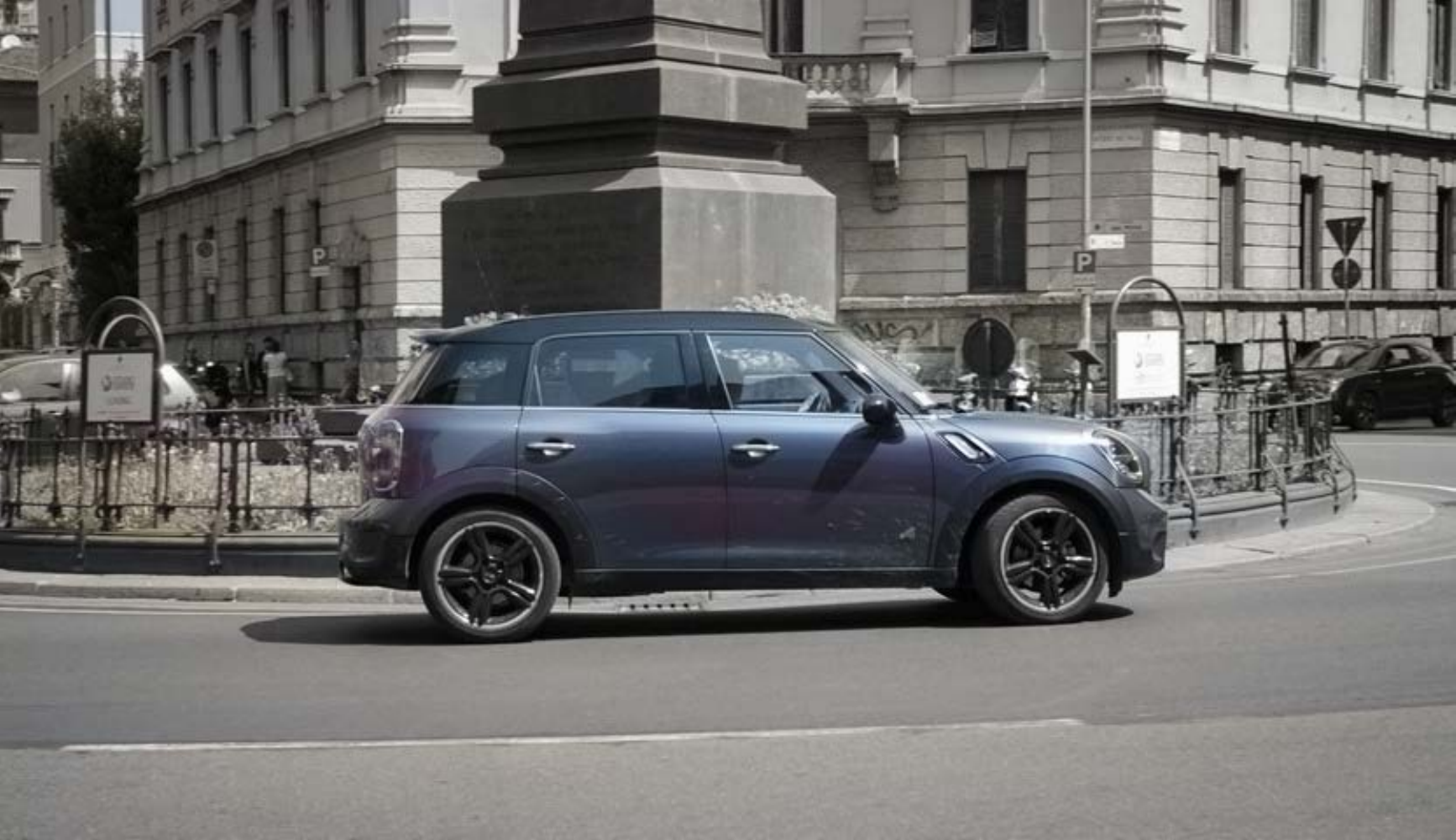}
	\rotatebox{90}{\;\footnotesize 1.68e-1}
	\\
	\rotatebox{90}{\quad\footnotesize Ours}
	
	&\includegraphics[width=0.105\textwidth]{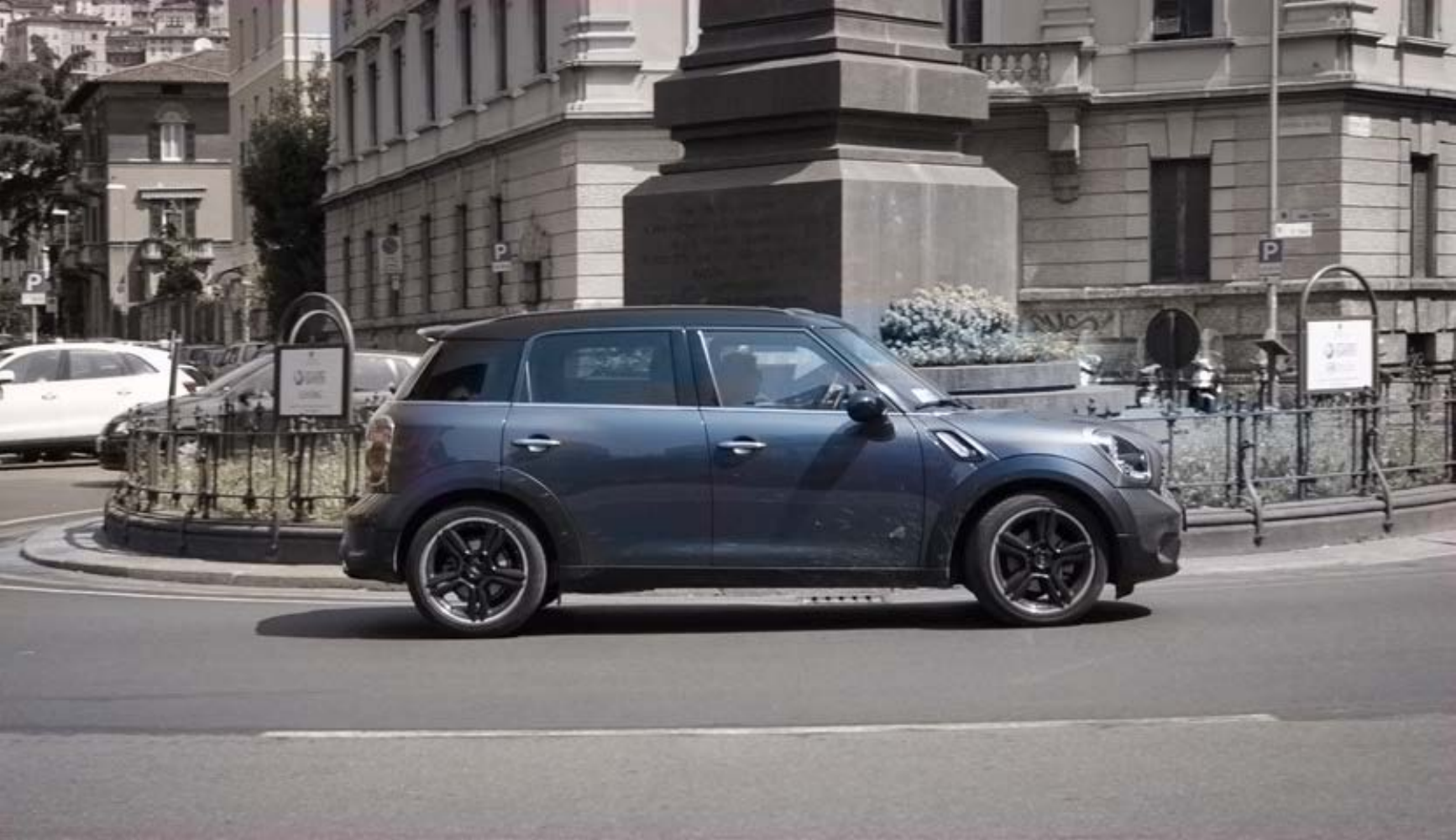}
	&\includegraphics[width=0.105\textwidth]{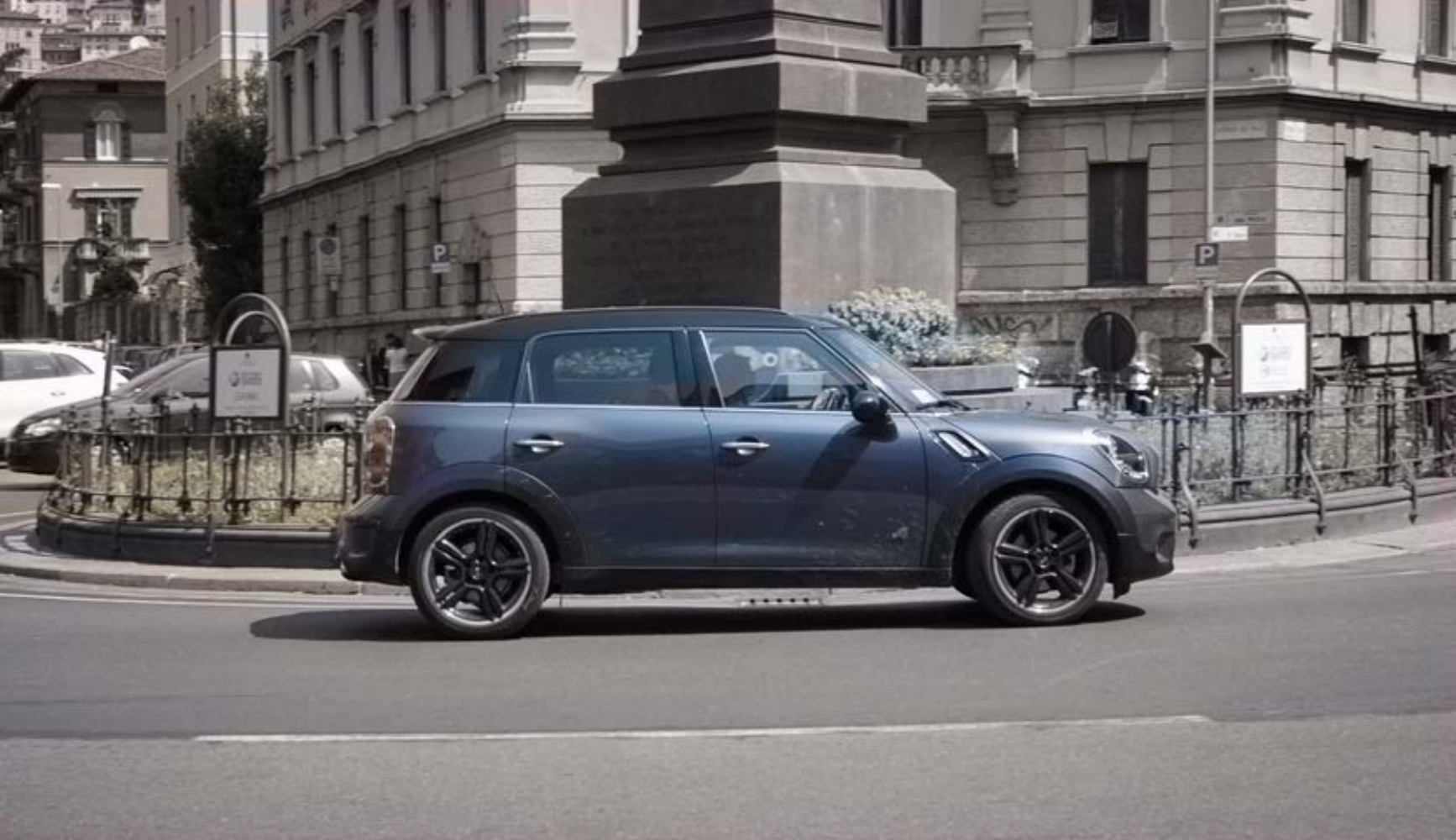}
	&\includegraphics[width=0.105\textwidth]{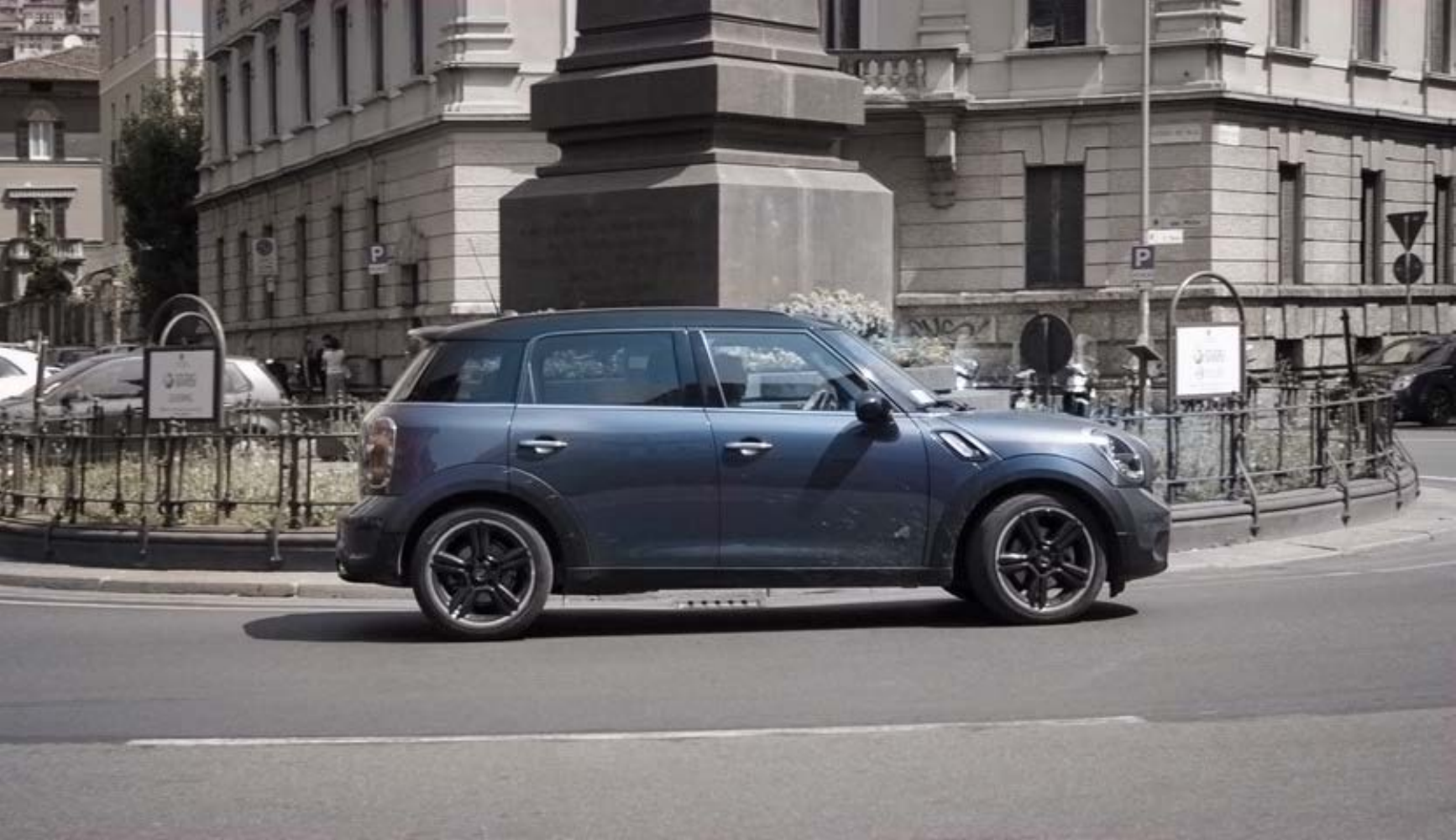}
	&\includegraphics[width=0.105\textwidth]{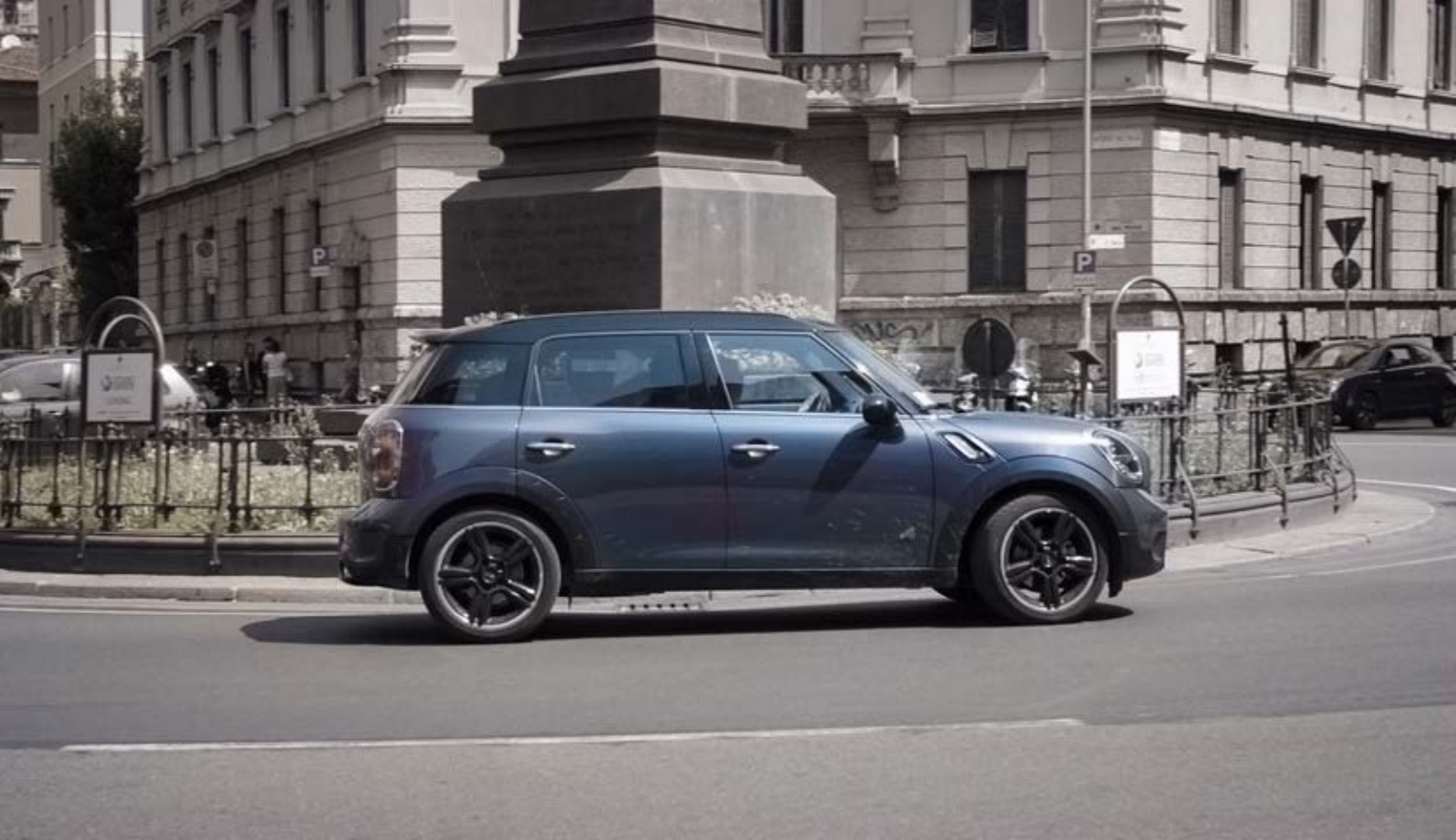}
	\rotatebox{90}{\;\footnotesize 1.61e-1}
	\\
	
	&\footnotesize frame 26  & \footnotesize frame 28 & \footnotesize frame 30 & \footnotesize frame 32\\ 
\end{tabular}
\caption{Comparison under warp error. We denote the corresponding warp error among the depicted frames in the right. We achieve the best results in both perception and statistics. }
\label{fig:warp_error} 
\end{figure}

\subsection{Ablation Studies}

\textbf{Illustrating Different Connection Schema.} 
We firstly conduct ablation studies on connection schema as is depicted in Tab.~\ref{tab:aba_input}. Detailed structure of each schema is shown in Fig.~\ref{fig:aba_input} where (c) is employed in our ST-HVC. 
The PSNR scores and drop rates from experiments demonstrate that temporal sharpness contributes to the reconstruction process and the direct concatenate operation is less effective than multiplication.

\textbf{Effectiveness of Different Modules.} 
We perform ablation studies on key components (i.e., the histogram guidance module, flow feature integration, temporal and spatial attention.) in the Videvo dataset. The statistical results are summarized in Tab.~\ref{tab:quantitative} and Tab.~\ref{tab:aba_components}. 
Without histogram, the proposed model still outperform other methods, and in many cases is merely second to the full model. 
Besides, the drop rates of PSNR are 6.42\% and 2.37\%, which demonstrate the great improvement brought by the histogram exploitation schema. 
The warp error grows nearly a half without flow, meaning the significance of flow integration paradigm.

Without temporal attention, warp error increases over a half, indicating the contribution of temporal branch in reducing the frames inconsistency.
Without spatial attention, the pixel-level scores drop heavily, showing that the spatial branch can better handle the texture information.

\textbf{Effectiveness of Flow Integration. }
In Fig.~\ref{fig:aba_flow_integration}, we demonstrate how will flow integration in JFHM affect the visual result. The first column shows the overall frame of dogs jumping and the following three figures show the details of green and red rectangles area while frames move forward.
Without flow integration, undersaturated area occurs. It is interesting to find that the area are most obvious in the first frame of one batch and will diminish along the frame sequences.
We argue that the phenomenon is stemmed from the inherent attributes of optical flow. With the flow integration, the undersaturated area vanishes and color becomes plausible and vivid again.

\subsection{ Comparing with State-of-the-Arts}
We conduct thorough experiments between our ST-HVC and several state-of-the-art approaches: image-based methods (i.e., CIC~\cite{zhang2016colorful}, IDC~\cite{zhang2017real}, InstColor~\cite{su2020instance} and GCP~\cite{wu2021towards}) and video-based methods (i.e., FAVC ~\cite{lei2019fully}, TCVC~\cite{liu2021temporally}. Moreover, we apply the blind temporal consistency~\cite{lai2018learning} on CIC~\cite{zhang2016colorful} and IDC~\cite{zhang2017real} to form another two groups of comparison.

\textbf{Quantitative Comparison.} 
We conduct quantitative comparisons on DAVIS and Videvo datasets where the results are summarized in Tab.~\ref{tab:quantitative}. 
The top four rows show the performance of image colorization methods and middle four rows present the result of video-based models.
In general, image-based methods can achieve relatively higher PSNR and SSIM scores, but their temporal coherence is poor. GCP ranks the first among image-based methods regarding warp error, since it utilizes generative color prior and the adjacent frames may have similar result when put into GAN encoder.
Video-based approaches like BTC is prone to boost the temporal consistency, consequently the warp error is improved. But the cost is singe image colorization quality when compared to original results of CIC and IDC. TCVC strike a balance between temporal consistency and colorization performance due to its bidirectional propagation of frame-level features.

As can be noticed in the bottom row of Tab.~\ref{tab:quantitative}, our method achieved the best score in most cases.
Due to our designed architecture, we rank the first in PSNR, SSIM and L2 error in both datasets. It demonstrates that the proposed method can generate the best textual and pixel-wise result closest to the ground truths. Besides, thanks to our flow integration, we outshine most models regarding the temporal consistency.

{\bf Comparison in Color.}
We conduct comparison regarding the chroma with representing and state-of-the-art approaches in Fig.~\ref{fig:compare_color}. 
From top to bottom, the figure demonstrates outstanding performance of the proposed method regarding the front color, background color, blurred pixels and the detail. Thanks to designed reference exploitation schema in JFHM, our proposed method can exploit the indication of color distribution. We can surprisingly assign the correct color in details as shown in the mouth of black-swan in bottom row.

{\bf Temporal Consistency Comparison.}
In Fig.~\ref{fig:warp_error}, we show the temporal consistency performance of our model.
Severe temporal flickering with inconsistent colors is occurred in InstColor.
As for GCP and TCVC, the overall results are great, but the color change of the body of cars between frames shows that they are still suffering from temporal inconsistency. Our proposed ST-HVC can generate satisfactory images with the least flickering and the lowest warp loss as is depicted in the figure.

\section{Conclusion}
We proposed a novel network with spatial-temporal connection schema to tackle the chroma assignment and temporal coefficient challenges in video colorization. 
To demonstrate its effectiveness, we conduct comparisons with the several state-of-the-art image and video-based methods. The experiment results show that the developed method achieves excellent scores on four quality metrics in two classic datasets. 

{\bf Limitations}
Despite the competitive performance, the proposed network cannot colorize well if scene changes rapidly, since  histogram of the middle frame is no longer representative. 
Future work will take these cases into account.

%
%

\end{document}